\documentclass[10pt,twocolumn,letterpaper]{article}

\usepackage{iccv}
\usepackage{times}
\usepackage{epsfig}
\usepackage{graphicx}
\usepackage{amsmath}
\usepackage{amssymb}
\usepackage{enumitem}
\usepackage{multicol}
\usepackage{float}
\usepackage{subfig}
\usepackage{pifont}

\usepackage{array}
\usepackage{times}
\usepackage{epsfig}
\usepackage{graphicx}
\usepackage{float}
\usepackage{wrapfig}
\usepackage{amsmath,amssymb,amsthm}
\usepackage{algorithm,algorithmicx,algpseudocode}
\usepackage{bm,xspace}
\usepackage{comment}
\usepackage{multirow}
\usepackage{balance}
\usepackage{url}
\usepackage{booktabs}
\usepackage{etoolbox,siunitx}
\usepackage{calc}
\usepackage{pifont,hologo}
\usepackage{color}
\usepackage{adjustbox}
\usepackage[normalem]{ulem}  
\usepackage[table]{xcolor}
\usepackage{nicefrac}
\usepackage[pagebackref,breaklinks,colorlinks,linkcolor=red,citecolor=blue,bookmarks=false]{hyperref}

\definecolor{lightgreen}{HTML}{D8ECD1}
\definecolor{lightorange}{HTML}{FFE4C4}
\definecolor{lightgray}{HTML}{DCDCDC}

\newcommand{\better}[1]{\colorbox{lightgreen}{#1}}

\newcommand{\worse}[1]{\colorbox{lightgray}{#1}}

\definecolor{bar}{HTML}{555555}

\definecolor{blue}{HTML}{0055cc}
\definecolor{red}{HTML}{cc1100}
\definecolor{orange}{HTML}{cc7700}
\definecolor{gray}{HTML}{efefef}
\definecolor{darkgreen}{rgb}{0.13, 0.55, 0.13}
\definecolor{darkgray}{HTML}{757575}

\renewcommand{\eqref}[1]{Eq.~\ref{#1}}

\newcolumntype{x}[1]{>{\centering\arraybackslash}p{#1}}
\newcolumntype{y}[1]{>{\raggedright\arraybackslash}p{#1}}
\newcolumntype{z}[1]{>{\raggedleft\arraybackslash}p{#1}}
\newcommand{\tablestyle}[2]{\setlength{\tabcolsep}{#1}\renewcommand{\arraystretch}{#2}\centering\footnotesize}

\DeclareMathSymbol{@}{\mathord}{letters}{"3B}

\newcommand\mypara[1]{\vspace{1mm}\noindent\textbf{#1}}

\makeatletter
\DeclareRobustCommand\onedot{\futurelet\@let@token\@onedot}
\def\@onedot{\ifx\@let@token.\else.\null\fi\xspace}
 
\def\ie{i.e\onedot}

\newcommand*{\Rom}[1]{\expandafter\@slowromancap\romannumeral #1@}
\newcommand*{\rom}[1]{\expandafter\romannumeral #1}

\def\1{\bm{1}}

\def\vf{{\bm{f}}}
\def\vg{{\bm{g}}}

\def\vp{{\bm{p}}}

\def\vv{{\bm{v}}}

\def\mF{{\bm{F}}}

\def\mI{{\bm{I}}}
\def\mJ{{\bm{J}}}

\def\mQ{{\bm{Q}}}

\DeclareMathAlphabet{\mathsfit}{\encodingdefault}{\sfdefault}{m}{sl}
\SetMathAlphabet{\mathsfit}{bold}{\encodingdefault}{\sfdefault}{bx}{n}

\def\sP{{\mathcal{P}}}

\def\ie{\emph{i.e.}}

\newcommand{\betterthanleft}[1]{\tiny{\textcolor{darkgreen}{(#1)}}}
\newcommand{\worsethanleft}[1]{\tiny{\textcolor{darkgray}{(#1)}}}

\newcommand{\authorskip}{\hspace{4mm}}

\iccvfinalcopy 

\begin{document}

\title{OCBEV: Object-Centric BEV Transformer for Multi-View 3D Object Detection}

\author{
Zhangyang Qi$^{1}$\footnotemark[2] \authorskip 
Jiaqi Wang$^{2}$\footnotemark[1] \authorskip
Xiaoyang Wu$^{1}$ \authorskip 
Hengshuang Zhao$^{1}$\footnotemark[1] \\
$^{1}$The University of Hong Kong, \quad $^{2}$Shanghai Artificial Intelligence Laboratory \\
{\tt\small \{zyqi, xywu3, hszhao\}@cs.hku.hk, wangjiaqi@pjlab.org.cn}
}

\maketitle
\renewcommand{\thefootnote}{\fnsymbol{footnote}}
\footnotetext[1]{Corresponding Authors.} 
\footnotetext[2]{The work is done during an internship at Shanghai AI Lab.}

\begin{abstract}
Multi-view 3D object detection is becoming popular in autonomous driving due to its high effectiveness and low cost. Most of the current state-of-the-art detectors follow the query-based bird's-eye-view (BEV) paradigm, which benefits from both BEV's strong perception power and end-to-end pipeline. Despite achieving substantial progress, existing works model objects via globally leveraging temporal and spatial information of BEV features,  resulting in problems when handling the challenging complex and dynamic autonomous driving scenarios. In this paper, we proposed an Object-Centric query-BEV detector OCBEV, which can carve the temporal and spatial cues of moving targets more effectively. OCBEV comprises three designs: \textbf{1) Object Aligned Temporal Fusion} aligns the BEV feature based on ego-motion and estimated current locations of moving objects, leading to a precise instance-level feature fusion. \textbf{2) Object Focused Multi-View Sampling} samples more 3D features from an adaptive local height ranges of objects for each scene to enrich foreground information. \textbf{3) Object Informed Query Enhancement} replaces part of pre-defined decoder queries in common DETR-style decoders with positional features of objects on high-confidence locations, introducing more direct object positional priors. Extensive experimental evaluations are conducted on the challenging nuScenes dataset. Our approach achieves a state-of-the-art result, surpassing the traditional BEVFormer by 1.5 NDS points. Moreover, we have a faster convergence speed and only need half of the training iterations to get comparable performance, which further demonstrates its effectiveness.
    
\end{abstract}

\vspace{-5pt}
\section{Introduction}
\label{sec:introduction}
\begin{figure}
    \begin{center}
    \includegraphics[width=1.0\linewidth]{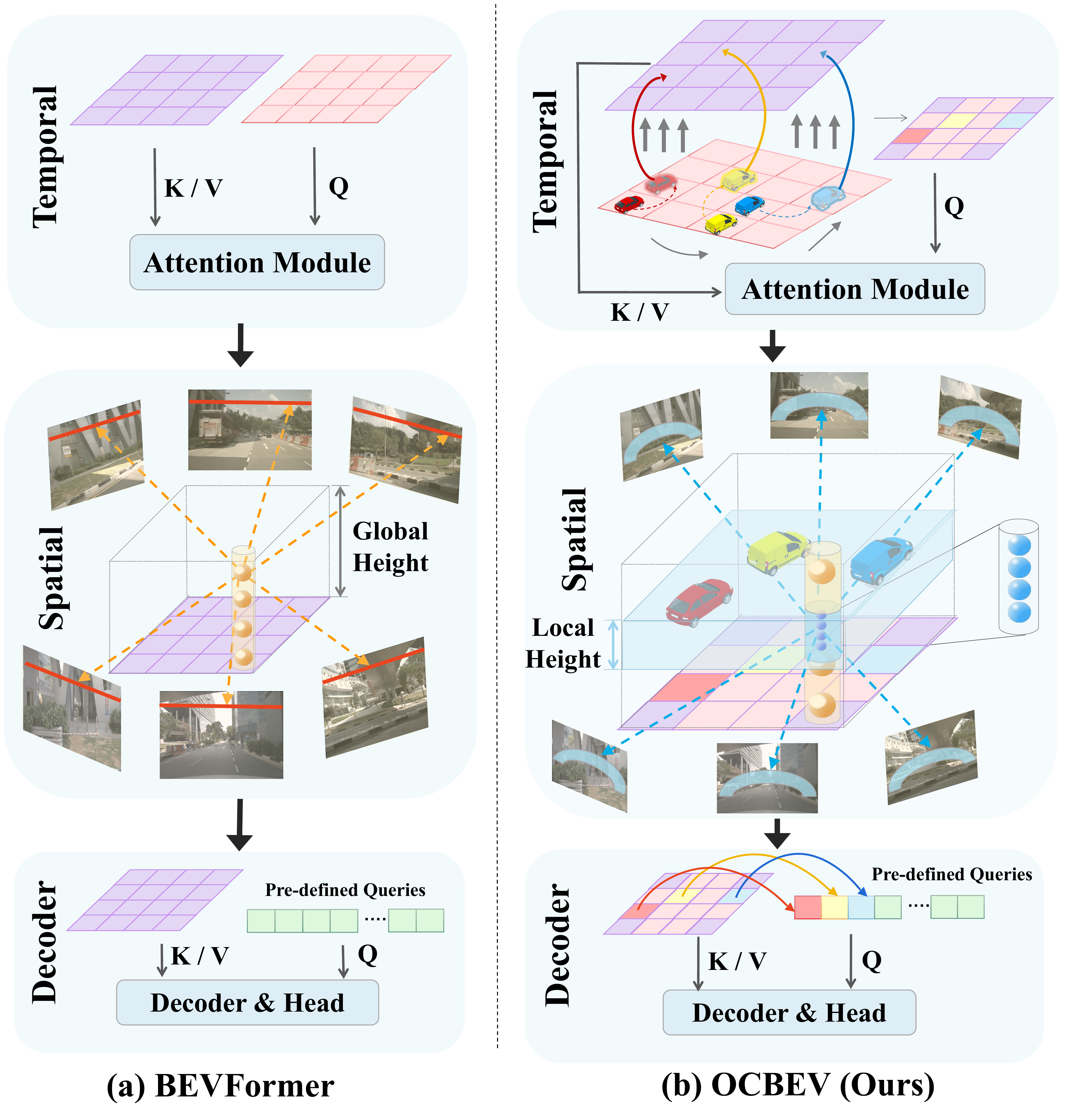}
    \end{center}
    \vspace{-15pt}
    \caption{
    \textbf{Comparison between (a) BEVFormer and (b) our OCBEV.} The top shows the temporal modeling. BEVFormer fuses BEV features in two timestamps via global attention while we fuse them more precisely via ego-motion and object-motion. The middle part is the spatial extraction from multi-view image features. BEVFormer only samples points in the global height range evenly while we conduct it densely within an adaptive local height range with most objects for each scene. The bottom shows the decoder. BEVFormer adopts pre-defined queries, whereas we introduce object location priors to the decoder via replacing some queries with BEV features on high-confidence locations.
    }
    \label{f1_teaser}
    \vspace{-15pt}
\end{figure}

Multi-view 3D object detection is attracting increasing attention in autonomous driving. Compared to monocular detectors~\cite{zhou2019objects, simonelli2019disentangling, wang2021fcos3d, wang2022probabilistic}, which lack depth information and have a limited field of view, multi-view detectors~\cite{wang2022detr3d, li2022bevdepth, li2022bevformer} provide a 360-degree view to fully obtain surrounding information. And unlike lidar-based methods~\cite{vora2020pointpainting, lang2019pointpillars, zhou2018voxelnet}, multi-view detectors have lower cost and rich semantic information such as traffic lights and lanes.

The multi-view detectors can be divided into \textit{object-query-based methods} ~\cite{wang2022detr3d, luo2022detr4d, lin2022sparse4d} and \textit{bird's-eye-view (BEV)-based solutions} ~\cite{huang2021bevdet, li2022bevdepth, li2022bevformer}. The transformer series, like DETR3D~\cite{wang2022detr3d}, adopt a DETR-style~\cite{carion2020end} network, where one query is assigned to one object in 3D space. However, since objects are not densely distributed, matching process become difficult, leading to low representation power. BEV-based methods employ the bird's-eye view (BEV) as an intermediate state to model the positions of objects and can be applied to other tasks like lane segmentation.

BEV-based methods can be further categorized into \textit{depth-BEV} ~\cite{philion2020lift, huang2021bevdet, li2022bevdepth} and \textit{query-BEV methods} ~\cite{li2022bevformer, yang2022bevformer}, based on the way to construct BEV feature. Depth-BEV methods predict a latent depth distribution to build frusta, while the other use pre-defined BEV queries to encode information from image features. Query-BEV methods like BEVFormer~\cite{li2022bevformer} stand out again for their temporal and spatial encoding capabilities. However, the current temporal modeling is done globally, overlooking the fact that the scene is dynamic and objects are moving. For spatial exploitation, 3D reference points in the global height range are used for projection as in Fig.~\ref{f1_teaser} (a), while this method falls short since the locations of moving objects is in a local height range area. Lastly, queries are all pre-defined for the decoder. It is known that object detection in 2D is difficult to optimize due to the scene's sparsity, thus the matching process for queries and objects is even more challenging for 3D multi-view scenes.

We build an \textit{\textbf{object-centric query-based bird's-eye view (BEV) detector}} that incorporates the benefits of both object-query methods and query-BEV methods. \textit{Frist}, to conduct temporal modeling, the query-BEV detector BEVFormer~\cite{li2022bevformer} uses deformable attention between historical and current BEV queries, as in Fig.~\ref{f1_teaser}(a) top. We propose a module called \textit{\textbf{Object Aligned Temporal Fusion}} that predicts the current locations of objects by using their historical locations and velocity. This allows for better feature representation when fused with current BEV features, as shown in the top of Fig.~\ref{f1_teaser}(b). \textit{Second}, to execute spatial exploitation, BEVFormer~\cite{li2022bevformer} only uses orange 3D reference points evenly distributed to the global height range, as in Fig.~\ref{f1_teaser}(a) middle. The top point hits the top of the image features like the orange line in the image features, which belong to the background. Contrarily, we develop \textit{\textbf{Object Focused Multi-View Sampling}} module that predicts a local height for each scenario adaptively where most moving objects are located at this height, and we densely sample 3D points in this area to ensure more of the points hit objects in the 2D image feature like the blue bar area in the image feature, as demonstrated in the middle of Fig.~\ref{f1_teaser}(b). \textit{Third}, the decoder of BEVFormer~\cite{li2022bevformer} uses pre-defined queries as in Fig.~\ref{f1_teaser}(a) bottom, which makes the object matching queries and objects more challenging. On the contrary, we design an \textit{\textbf{Object Informed Query Enhancement}} module to replace part of the decoder queries with high-confidence locations of objects via a centerness heatmap supervision head after the encoder, as illustrated in the bottom of Fig.~\ref{f1_teaser}(b) which leads to a faster convergence speed for the whole network. 

We name the whole algorithm as Object-Centric query-BEV detector OCBEV, and conduct extensive experimental evaluations on the challenging nuScenes~\cite{caesar2020nuscenes} benchmark to validate its effectiveness. The proposed OCBEV achieves a state-of-the-art result, surpassing the classical BEVFormer by 1.5 NDS point. Further, it converges faster and only needs half of the training iterations to get comparable performance. Our code and models will be made publicly available. In summary, our contributions are as follows:
\vspace{5pt}

\begin{itemize}[topsep=0pt, parsep=1pt, itemsep=0pt, partopsep=0pt]
\item We find that object-centric modeling is the key to the query-BEV 3D detectors. Existing methods only consider extracting temporal and spatial aspects globally, whereas we conduct in-depth design to overcome it.
\item We present a powerful BEV detector named OCBEV, which consists of three newly proposed novel modules that can better conduct temporal modeling, execute spatial exploitation, and construct strong decoders.
\item We evaluate our method on the challenging nuScenes dataset. Our OCBEV method can achieve state-of-the-art results in the vision-based methods. And our network converges faster and cut training iterations in half to get comparable results with existing methods.
\end{itemize}

\section{Related Work}
\label{sec:related_work}
\noindent\textbf{Multi-view vision-based 3D detection.}
For multi-view camera-only object detection, various object-query-based algorithms have been developed. DETR3D~\cite{wang2022detr3d} follows the 2D detector DETR~\cite{carion2020end} and replaces the 2D queries with the 3D queries. PETR~\cite{carion2020end} develops position embedding transformation. Polarformer~\cite{jiang2022polarformer} and Polar-DETR~\cite{chen2022polar} advocate exploiting the Polar coordinate system. Sparse4D~\cite{lin2022sparse4d} uses hierarchy feature fusion for sparse 3D detection.

Bird’s-eye-view (BEV) is a new trend for multi-view perception tasks. BEV is not originally from 3D detection tasks, and early works consider transforming perspective features to BEV features~\cite{roddick2018orthographic, pan2020cross} like the pseudo-lidar~\cite{wang2019pseudo, weng2019monocular}. Current BEV methods mainly extend to the 360-degree sensor to build the surrounding environment~\cite{philion2020lift, can2021structured, zhou2022cross, zhang2022beverse}. LSS~\cite{philion2020lift} predicts an implicit depth distribution to build the frusta from six cameras and construct the BEV feature by splatting frusta. While CVT~\cite{zhou2022cross} bridges perspective-view and BEV-view features by a camera-aware positional encoding module and dense cross-attention.

BEV is widely used in 3D object detection, which is originally a kind of lidar-based representation~\cite{harley2022simple, mohapatra2021bevdetnet}. TiG-BEV~\cite{huang2022tig} and CMT~\cite{yan2023cross} both use BEV feature to combine lidar-based methods and vision-based methods. For vision-based methods only, according to the way to build the BEV feature, these methods can be divided into depth-BEV-based methods~\cite{huang2021bevdet, li2022bevstereo, li2022bevdepth, wang2022mv} and query-BEV-based methods~\cite{li2022bevformer, yang2022bevformer}. For depth-BEV-based methods, BEVDet~\cite{huang2021bevdet} gives the paradigm of BEV 3D detection from LSS~\cite{philion2020lift}. BEVstereo~\cite{li2022bevstereo} and Bevdepth~\cite{li2022bevdepth} extend it to stereo-view input and depth supervision, respectively. Query-BEV-based methods like BEVFormer~\cite{li2022bevformer} do not need to build the latent depth but extract information from the perspective-view feature directly. Therefore, we focus on designing a query-BEV-based 3D detector. More details of \textit{\textbf{object-query-based}}, \textit{\textbf{BEV-query-based}}, and \textit{\textbf{BEV-depth-based methods}} will be explained in the Appendix~\ref{sec:sup_A1_3_types_detection}.

\noindent\textbf{Temporal and spatial exploration in 3D detection.}
Temporal modeling and spatial exploitation are extremly important in BEV detection. The former means modeling the history information since autonomous driving can be seen as a continuous video and history will give clues to the current frame. Acutually, many 3D detectors have considered the temporal modeling~\cite{li2022bevdepth, li2022bevstereo, li2022bevformer}. There are also some works that only utilize temporal information. BEVDet4d~\cite{huang2022bevdet4d}, DETR4D~\cite{luo2022detr4d} and PETR v2~\cite{liu2022petrv2} all do temporal modeling based on BEVDet~\cite{huang2021bevdet}, DETR3D~\cite{wang2022detr3d}, and PETR~\cite{liu2022petr}, respectively. Some works only study the effect of the temporal information, TWW~\cite{park2022time} uses long and short temporal image feature fusion to get the BEV feature. However, all of them only consider BEV temporal modeling at the global level while we consider aligning moving objects of their displacement.
Spatial information is to make the BEV feature extract information from the perspective view. And as we mentioned before, both depth-based~\cite{huang2021bevdet, li2022bevstereo, li2022bevdepth, wang2022mv} and query-based methods~\cite{li2022bevformer, yang2022bevformer} try to bridge these two views. But they meet the same problem, \ie, the spatial exploitation only bridges in a global range while most objects are densely in a local range. Some works like BEV-SAN~\cite{chi2022bev} and OA-BEV~\cite{chu2023oa} also try to pay more attention to these moving objects. We propose to predicts an adaptive local height range for each scenario where most moving objects are located at this height range for better feature learning.

\noindent\textbf{Decoder enhancement in object detection.}
Traditional CNN-based two-stage detectors~\cite{girshick2014rich, girshick2015fast, lin2017feature} first give rough proposals,followed by precise predictions. In DETR-based methods~\cite{carion2020end, zhu2020deformable, yao2021efficient, li2022dn, zhang2023dino}, the encoder-decoder structure can be regarded as a two-stage pipeline. Two-stage deformable DETR~\cite{zhu2020deformable} feeds the high score encoder embeddings to the decoder. Efficient DETR~\cite{yao2021efficient} enhances decoder queries by selecting top-k positions from the encoder’s output. Apart from giving positional clues to the decoder, DN-DETR~\cite{li2022dn} and DINO~\cite{zhang2023dino} add noisy labels and bounding boxes to the decoder for speeding up. In query-based 3D object detection, BEVFormer v2~\cite{yang2022bevformer} adopts a perspective view head after the image features to provide reference points and positional embeddings to the decoder. While we only need to provide positional instead of semantic information, so we add a centerness heatmap supervision which not only gives the decoder enough positional hints but makes the network end-to-end trainable as well. More details about decoders will be shown in the Appendix~\ref{sec:sup_A2_3_DETR_Decoder}.

\section{Object-Centric BEV Transformer}
\label{sec:methodlogy}
\begin{figure*}[t]
    \centering
    \includegraphics[width=0.9\linewidth]{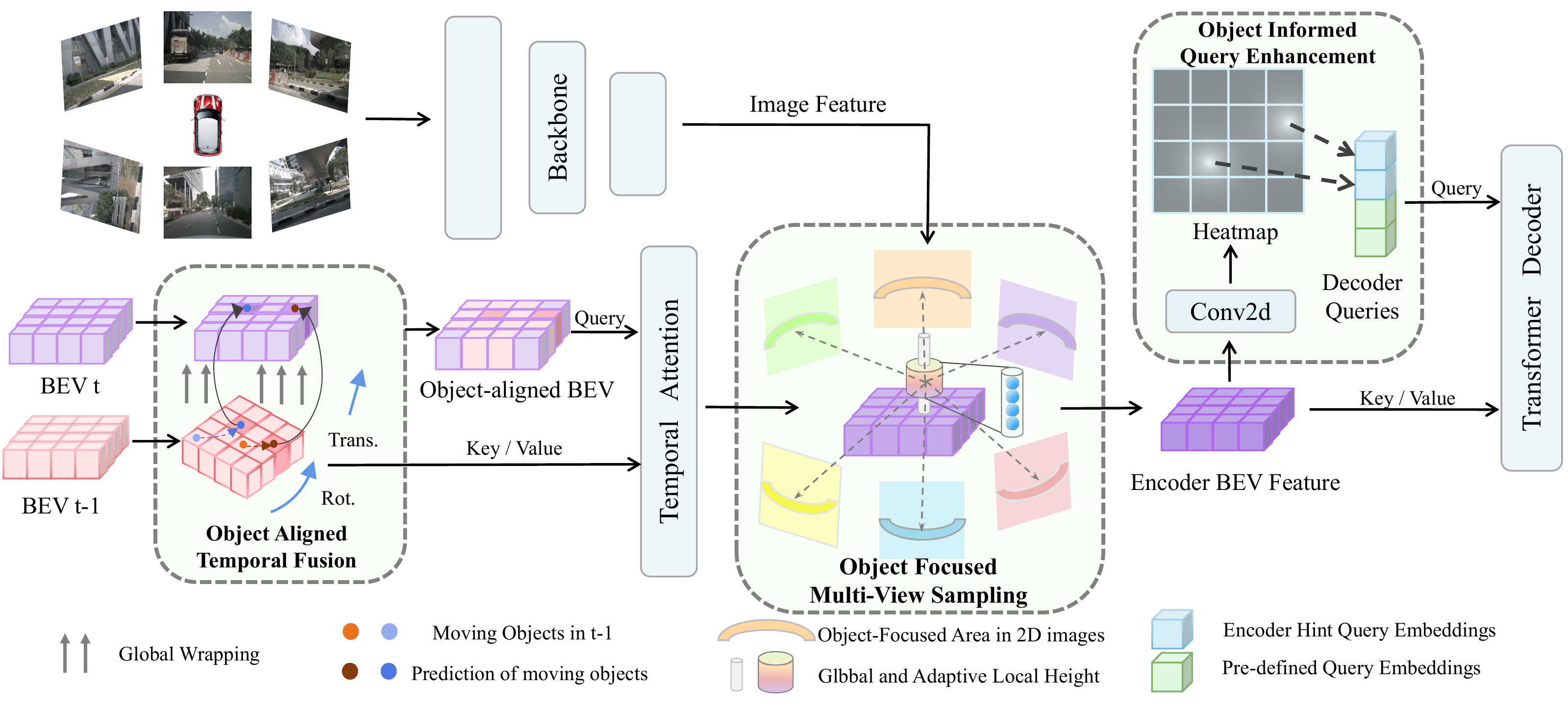}
    \vspace{-1mm}
    \caption{\textbf{Overall architecture of OCBEV.} \textit{(a) Object Aligned Temporal Fusion}: warps BEV $t-1$ to BEV $t$ by ego motion transformation, then fuses moving objects by predicting their current locations via their velocity. \textit{(b) Object Focused Multi-View Sampling}: pre-defines points in the 3D space and then projects them to perspective-view image features. Apart from sampling points in the global height range, we propose an adaptive local object-dense height with most objects for each autonomous driving scene and sample points densely in it. \textit{(c) Object Informed Query Enhancement}: A heatmap supervision is added after the encoder output and replaces part of pre-defined decoder queries with high-confidence locations of objects.}
    \label{f2_architecture}
    \vspace{-2mm}
\end{figure*}
\subsection{Overall Architecture}

The overall framework of our OCBEV is illustrated in Fig.~\ref{f2_architecture}. As most query-BEV methods like BEVFormer~\cite{li2022bevformer} and BEVFormer v2~\cite{yang2022bevformer}, we start by encoding the pre-defined BEV queries with temporal modeling over historical BEV queries and spatial exploitation on perspective-view image features. Sec.~\ref{sec:Object Aligned Temporal Fusion} introduces Object Aligned Temporal Fusion, which aligns and fuses historical BEV, considering not only the background but also moving objects. And it can be divided into \textit{ego-motion temporal fusion} and \textit{object-motion temporal fusion} in consideration of ego-motion and object-motion, respectively. Besides, we utilize attention between these two BEV-feature timestamps after the fusion. For spatial information, we introduce Object Focused Multi-View Sampling in Sec.~\ref{sec:Object Focused Multi-View Sampling}, which is based on spatial-cross attention in the BEVFormer~\cite{li2022bevformer}. We predict an adaptive local height within an object-focused area and sample the corresponding dense 3D points. Afterward, we project them to 2D image features and form a bar area where moving objects are densely distributed. After encoding, we construct Object Informed Query Enhancement for the decoder in Sec.~\ref{sec:Object Informed Query Enhancement}, which includes a heatmap head that predicts the object centerness, and exchanges some pre-defined decoder queries with high-confidence locations that give positional hints. Finally, a detection head like CenterHead~\cite{yin2021center} is implemented to predict the final results.

\subsection{Object Aligned Temporal Fusion}
\label{sec:Object Aligned Temporal Fusion}
Recent BEV-based detectors incorporate temporal information by fusing the BEV features from previous timestamps, resulting in significant performance improvements. However, the existing modeling process meets problems when dealing with complex and dynamic autonomous driving scenarios and is incapable of discovering the dynamic characteristics. As an example, the depth-BEV series approaches utilize two timestamps as a stereo view to obtain the depth information, leaving the capturing of the moving status of objects untouched. Similarly, as representatives of query-BEV methods, BEVFormer~\cite{li2022bevformer} attempts to utilize deformable attention to capture history information, but also ignore the geometry modeling for moving objects.

Existing BEV temporal fusion strategies fail to address two critical problems: i) Fail to track the ego-motion information of the host vehicle, which is indispensable for modeling the driving scenario; ii) Fail to leverage geometry similarity, which is crucial to handle fast-moving vehicles. To tackle these challenges, we propose a novel solution Object Aligned Temporal Fusion. Specifically, we perform ego-motion temporal fusion and object-motion temporal fusion, resulting in a more comprehensive and effective fusion of temporal information.

\begin{figure}[t]
    \centering
    \includegraphics[width=0.99\linewidth]{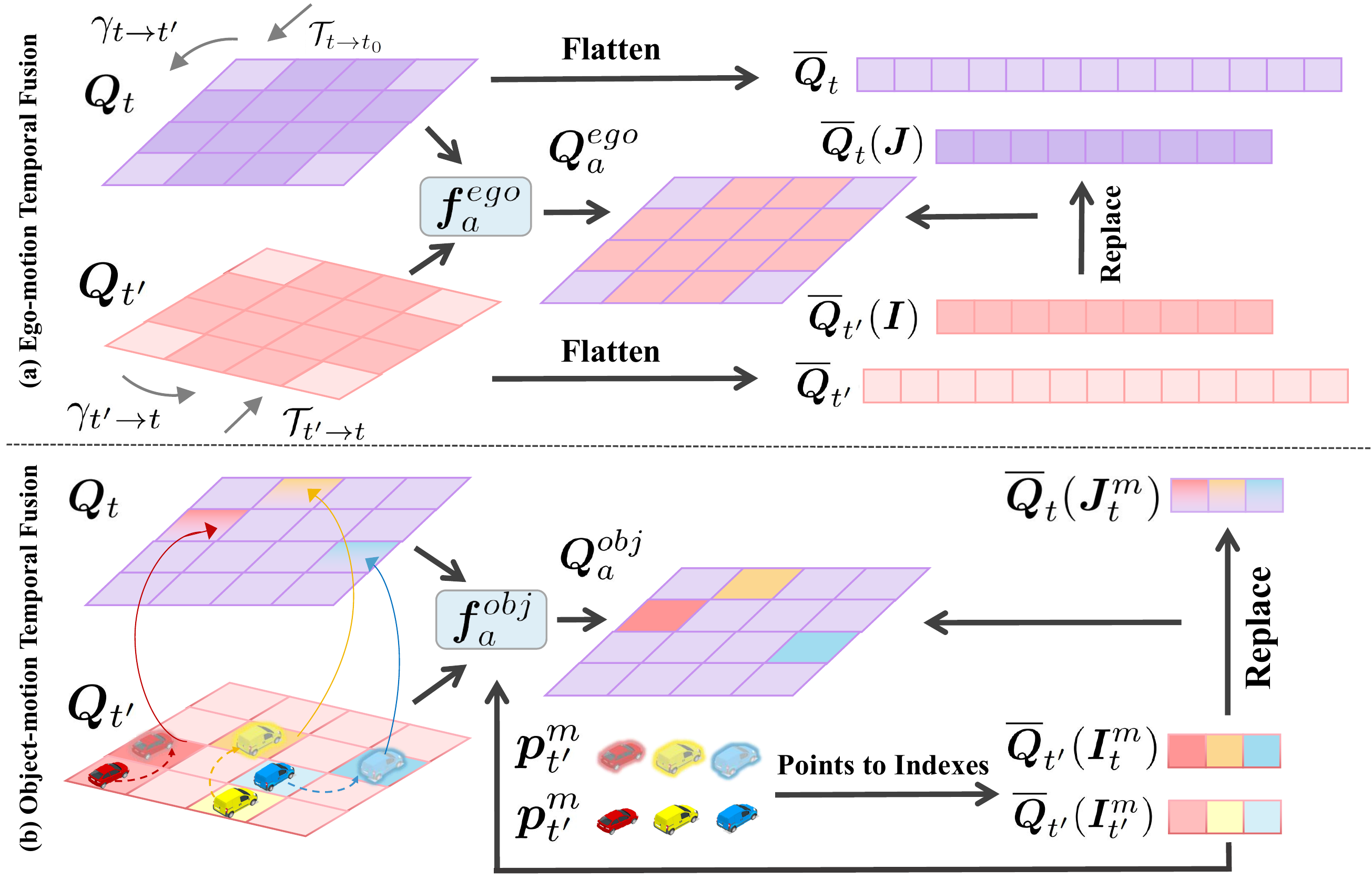}
        \caption{\textbf{Object Aligned Temporal Fusion.} (a) \textit{Ego-motion temporal fusion} finds overlapping indices of adjacent BEV features by the ego-motion globally. Then we flatten BEV queries and replace current embeddings with corresponding previous embeddings. Original embeddings are added for fusion. (b) \textit{Object-motion temporal fusion} is for moving objects. We predict their current locations and indices by velocity and use the above ego-motion fusion way to find and replace their corresponding current queries.}
    \label{f3_temporal}
    \vspace{-10pt}
\end{figure}

\mypara{Ego-motion temporal fusion.} BEV ego-motion temporal fusion differs from points alignment in 3D space. To fuse a current query feature with its corresponding feature from a previous timestamp, we have to perform a bidirectional transformation to determine the mapping relationship which ensures precise alignment and fusion of features between different timestamps. For convenience, we define the previous timestamp as $t^\prime$ and the current timestamp as $t$. We also denote the BEV queries as $\mQ_t\!\in\! \mathbb{R}^{C_{q}\!\times\!H_{q}\!\times\!W_{q}}$ and its flatten version as $\overline{\mQ}_t\!\in\! \mathbb{R}^{C_{q}\!\times\!(H_{q}W_{q})}$. Fig.~\ref{f3_temporal} (a) demonstrates the process of aligning ego-motion. We aim to determine the mapping $\sP = \{(i, j)\}_{t^\prime,t}$, where $\overline{\mQ}_{t^\prime}(i) \rightarrow \overline{\mQ}_{t}(j)$. The set $\sP$ contains $N_p$ pairs, indicating the number of overlapping queries. The indices of overlapping queries in timestamp $t^\prime$ and $t$ are denoted as $\mI$ and $\mJ \in \mathbb{R}^{1\times N_p}$, respectively.

The main idea to obtain $\mI$ is to transform $\mQ_{t^\prime}$ to the $t$ reference frame for selecting the queries overlapped with $\mQ_{t}$. We simplify the problem by assuming that vehicles are moving on a plane, which means that the $3\!\times\!3$ rotation matrix $\mathcal{R}_{t^\prime \rightarrow t}$ can be reduced to a simplified $\gamma_{t^\prime \rightarrow t}$ representing the yaw angle along the $Z$ axis. The $3\!\times\!1$ translation vector is denoted as $\mathcal{T}_{t^\prime \rightarrow t}$, and we only use the $xy$ translation because the motion is on a plane. The selected indices via the transformation from $\mQ_{t^\prime}$ to $\mQ_{t}$ are denoted as $\mI$ and can be written as Eq.~\ref{eqn:global_alignment}. Similarly, we obtain $\mJ$ as well.
\begin{equation}
\begin{aligned}
\label{eqn:global_alignment}
    \mI = [ \vg_{t} \big(  \vg_{r} ( \mQ_{t^\prime}, \gamma_{t^\prime \rightarrow t} ) , \mathcal{T}_{t^\prime \rightarrow t} \big) \in \mQ_t] \\
    \mJ = [ \vg_{t} \big(  \vg_{r} (\mQ_{t}, \gamma_{t \rightarrow t^\prime}) , \mathcal{T}_{t \rightarrow t^\prime} \big) \in \mQ_{t^\prime}]
\end{aligned}
\end{equation}
here $\vg_{t}$ represents translation operation, whereas $\vg_{r}$ represents rotation operation. For the BEV queries fusion, we replace overlapping BEV queries $\overline{\mQ}_{t}(\mJ)$ with corresponding previous BEV queries $\overline{\mQ}_{t^\prime}(\mI)$ as $\overline{\mQ}_{t^\prime}(\mI) \rightarrow \overline{\mQ}_{t}(\mJ)$, and add the replaced BEV queries and current BEV queries $\mQ_{t}$. All alignment and fusion operations are expressed as:
\begin{equation}
    \begin{aligned}
    \mQ_{a}^{ego} = \vf_{a}^{ego} \left( \mQ_{t^\prime}, \mQ_t \right) + \mQ_{t}
    \end{aligned}
\end{equation}
Here $\vf_{a}^{ego}$ is the function to align the BEV feature by ego-motion bidirectional transformation as mentioned above, we fuse it with current BEV embeddings $\mQ_{t}$ by adding them together and $\mQ_{a}^{ego}$ is the fused BEV queries.

\mypara{Object-motion temporal fusion.} Apart from the ego-motion temporal fusion, we also align and fuse moving objects as Fig.~\ref{f3_temporal} (b). Suppose there are $N_m$ objects in timestamp $t^\prime$. $\vp_{t^\prime}^{m} \!\in\! \mathbb{R}^{N_m \times 2}$ is their real $xy$ coordinates with the corresponding indices $\mI_{t^\prime}^{m}$ on the $\mQ_{t^\prime}$. We predict velocity $\vv_{t^\prime}^{m} \!\in\! \mathbb{R}^{N_m \times 2}$ to find the corresponding indexes $\mJ_{t}^{m}$ in $\mQ_t$ and replace them as
$\overline{\mQ}_{t^\prime}(\mI_{t^\prime}^{m}) \rightarrow \overline{\mQ}_{t}(\mJ_{t}^{m})$.
Specifically, we first predict the locations after the motion in $t$ reference frame as $\vp_{t}^{m} = \vp_{t^\prime}^{m} + \vv_{t^\prime}^{m}(t - t^\prime)$. Then convert coorinates into indices as $\mI_{t}^{m}$. To get its corresponding after-motion indices $\mJ_{t}^{m}$ in $\mQ_{t}$, we rotate and translate $\overline{\mQ}_{t^\prime}(\mI_{t}^{m})$ as:
\begin{equation}
    \begin{aligned}
    \mJ_{t}^{m} = [ \vg_{t} \Big( \vg_{r} \left(\overline{\mQ}_{t^\prime}(\mI_{t}^{m}), \gamma_{t^\prime \rightarrow t}\right), \mathcal{T}_{t \rightarrow t^\prime} \Big) \in \mQ_{t^\prime}]
    \end{aligned}
\end{equation}
$\mI_{t}^{m}$ and $\mJ_{t}^{m}$ form the object-level mapping $\overline{\mQ}_{t^\prime}(\mI_{t}^{m}) \rightarrow \overline{\mQ}_{t}(\mJ_{t}^{m})$. Having the mapping relationship, we align and fuse the moving objects as before. The whole process is: 
\begin{equation}
    \begin{aligned}
    \mQ_{a}^{obj} = \vf_{a}^{obj} \left( \mQ_{t^\prime}, \mQ_t, \vp_{t^\prime}^{m} \right)
    \end{aligned}
\end{equation}
Here $\vf_{a}^{obj}$ is the object-motion temporal alignment and fusion function locally between two timestamps. Apart from previous $\mQ_{t^\prime}$ and current $\mQ_t$ BEV queries, we have to input the original locations of objects $\vp_{t^\prime}^{m}$. Overall, the Object Aligned Temporal Fusion can be written as:
\begin{equation}
    \begin{aligned}
    \mQ_{a} &= \mQ_{a}^{ego} + \mQ_{a}^{obj}\\
    &= \vf_{a}^{ego} \left( \mQ_{t^\prime}, \mQ_t \right) + \vf_{a}^{obj} \left( \mQ_{t^\prime}, \mQ_t, \vp_{t^\prime}^{m} \right) + \mQ_{t}
    \end{aligned}
\end{equation}
where $\mQ_{a}$ is the aligned and fused BEV embeddings between $t^\prime$ and $t$ timestamps. This operation contains ego-motion and object-motion temporal fusion.

After the alignment and fusion, to fully utilize the historical information, we adopt the deformable DETR attention~\cite{zhu2020deformable} between aligned BEV feature $\mQ_{a}$ and previous BEV feature $\mQ_{t^\prime}$ as well. After finishing the temporal modeling, the BEV feature will be fed into the spatial module to extract information from perspective-view image features.

\subsection{Object Focused Multi-View Sampling}
\label{sec:Object Focused Multi-View Sampling}
Spatial exploitation for BEV queries is to extract information from perspective-view image features. All query-based methods depend on the deformable DETR~\cite{zhu2020deformable}, which makes the image features as keys and values. They pre-define 3D reference points for the BEV queries and project them to the 2D image features to sample keys and values. The question is whether the height of predefined 3D reference points is in the overall detection height range. Nevertheless, most vehicles are located at a local height. So our Object Focused Multi-View Sampling will predict an adaptive local height range that contains most objects with consideration of vehicles such as bus and truck and the height range can be adaptive for each unique scene.

We follow the spatial attention module based on the deformable attention~\cite{zhu2020deformable} in the BEVFormer~\cite{li2022bevformer}, which defines $N_h$ pillar-like reference points along the $z$-axis for one specific BEV query $\mQ(u, v)$. The reference points are $\vp_{3d}\!\in\! \mathbb{R}^{N_h \!\times\! 3}$. All reference points share the same $xy$ coordinates, which are the center of the BEV embedding corresponding area. Regarding the $z$ coordinate, these points are distributed along the $z$-axis in the range $\textbf{Z}_h$ evenly. So we denote $\vp_{3d}(\textbf{Z}_h)$ which means that the $z$ coordinate of reference points $\vp_{3d}$ are distributed evenly in the $\textbf{Z}_h$ height range. Thus the spatial attention is shown below:
\begin{equation}
\begin{aligned}
\label{spatial attention}
    \text{SpaA}(\mQ_t, \textbf{Z}_{h})
    &= \text{DAttn} \Big( \mQ_t, \mF_t, \vg_{p} \big( \vp_{3d} \left( \textbf{Z}_h \right) \big) \Big)
\end{aligned}
\end{equation}
where $\text{SpaA}$ is the spatial attention and inputs are BEV queries $\mQ_t$ and height range $\textbf{Z}_{h}$. $\text{DAttn}$ is the deformable attention~\cite{zhu2020deformable}. $\mQ_t$ is the queries, and multi-view image features $\mF_t$ are the keys and values. $\vg_{p}$ is the projection function for projecting 3D reference points to the 2D multi-view image features through extrinsic and intrinsic matrices. 

In autonomous driving scenarios, most objects, including cars, buses, and pedestrians, are in a local height range. For the nuScenes dataset~\cite{caesar2020nuscenes}, the detection global range $\textbf{Z}_{h,g}$ is from $-5m$ to $3m$. Apart from the global range, we notice that most moving objects are concentrated in the height range of $[-2m, 2m]$~\cite{chi2022bev} because the camera sensor is usually located on the top of the data collection vehicle, so we define the local height $\textbf{Z}_{h,l}$ from $-2m$ to $2m$. We also predict an adaptive offset $\Delta h$ for each scene to handle scenes with high vehicles like buses and trucks which is also common in autonomous driving scene. The adaptive local height is denoted as $\hat{\textbf{Z}}_{h,l} = \textbf{Z}_{h,l} + \Delta h$.

As shown in the middle right part of Fig.~\ref{f1_teaser} and Fig.~\ref{f2_architecture}, these 3D points in the local height range are projected to the 2D image features and form bar areas that contain dense objects. So we sample 3D reference points in not only the global height range but also densely in the local height range. In this way, the spatial attention inside our Object Focused Multi-View Sampling is shown below:
\begin{equation}
\small
\begin{aligned}
\label{OF-SpaA}
    \text{OFSpaA}(\mQ_t, \textbf{Z}_{h}) &= \text{SpaA}(\mQ_t, \textbf{Z}_{h,g}) + \text{SpaA}(\mQ_t, \hat{\textbf{Z}}_{h,l})
\end{aligned}
\end{equation}
Here, $\text{OFSpaA}$ means the spatial attention by Object Focused Multi-View Sampling, which is the sum of spatial attention with 3D reference points in global height range and adaptive local height range to focus the moving objects and make BEV's extraction power from image features stronger.
\begin{figure}[t]
    \centering
    \includegraphics[width=0.99\linewidth]{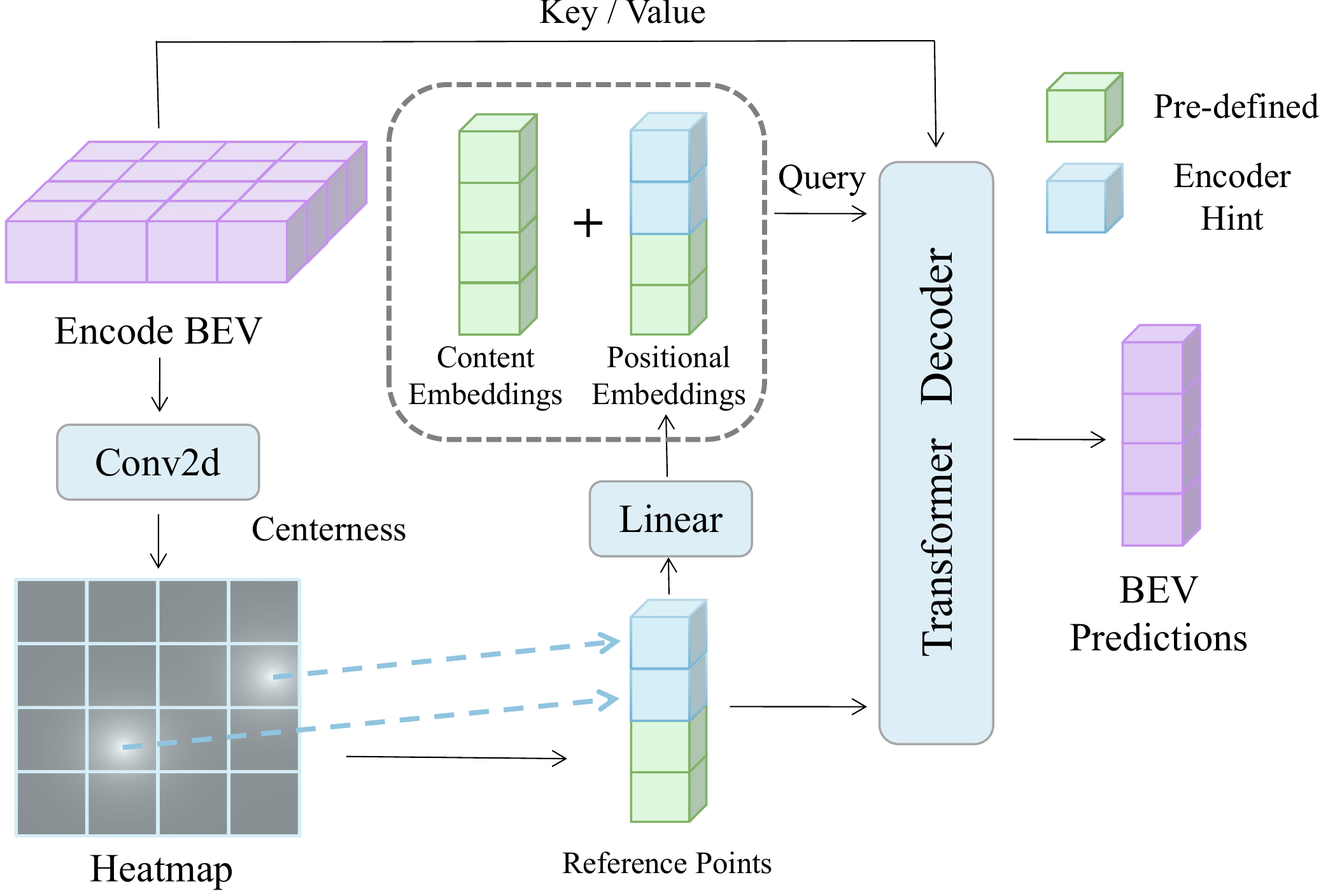}
    \vspace{-1mm}
    \caption{\textbf{Object Informed Query Enhancement.} Apart from feeding the BEV feature as keys and values, we use a heatmap supervision after the encoder BEV feature to get a confidence heatmap and replace the pre-defined reference points with high-confidence locations partially. The queries are divided into content embeddings that are predefined and positional embeddings that are from the reference points.}
    \label{f4_decoder}
    \vspace{-5mm}
\end{figure}

\subsection{Object Informed Query Enhancement}
\label{sec:Object Informed Query Enhancement}
As for the decoder, as mentioned above, DETR~\cite{carion2020end} style pre-defines queries randomly, which leads to difficulty in convergence and will be shared in all scenes. Inspired by two-stage CNN-based methods, we use encoder output to deliver a positional hint to the decoder. Our decoder is shown in Fig.~\ref{f4_decoder}, which divides the queries to content and positional embeddings following deformable DETR~\cite{zhu2020deformable}.

Unlike two-stage deformable DETR~\cite{zhu2020deformable} that directly adds a detection head after the encoder, we add a heatmap head based on the encoder output. Since the outdoor 3D scene is within a wide range, we can focus more on the position in the BEV plane, rather than other properties like box size, velocity, attribute, etc. Here we introduce object centerness as defined in the FCOS3D~\cite{wang2021fcos3d} and Centernet~\cite{zhou2019objects}. The centerness is represented by the 2D Gaussian distribution, which ranges from zero to one, and we utilize the binary cross entropy loss to conduct the training.

Apart from extra heatmap supervision, we replace some predefined queries with the BEV features on high-confidence locations. To be specific, for the decoder transformer, the keys and values are output features of the BEV encoder. We divide the queries into content and positional embeddings. The origin deformable DETR~\cite{zhu2020deformable} initializes content embeddings positional embeddings and reference points randomly. Whereas in our framework, we use BEV points with high-confidence scores as reference points to partially replace the pre-defined part of reference points and positional embeddings through a linear layer whose input is replaced reference points, which brings beneficial positional priors to make the network converge quickly. 

\section{Experiments}
\label{sec:experiments}

\begin{table*}[t!]
\begin{center}
    \tablestyle{6.3pt}{1.02}
    \begin{tabular}{c l c| c |c c| c c c c c c }
\toprule
Style & Method & Backbone & Epoch & NDS~$\uparrow$ & mAP~$\uparrow$  & mATE~$\downarrow$ & mASE~$\downarrow$ & mAOE~$\downarrow$ & mAVE~$\downarrow$ & mAAE~$\downarrow$    \\
\midrule

\multirow{3}*{Monocular} & CenterNet~\cite{zhou2019objects} & DLA~\cite{yu2018deep} & - & 0.328 & 0.306 & 0.716 & 0.264 & 0.609 & 1.426 & 0.658  \\

& FCOS3D~\cite{wang2021fcos3d} & ResNet-101 & - & 0.415 & 0.343 & 0.725 & 0.263 & 0.422 & 1.292 & \textbf{0.153}  \\

& PGD~\cite{wang2022probabilistic} & ResNet-101 & 24 & 0.428 & 0.369 & 0.683 & 0.260 & 0.439 & 1.268 & 0.185  \\

\midrule

\multirow{2}*{Depth-based} & BEVDet$^\dagger$~\cite{huang2021bevdet} & Swin-T & 24 & 0.417 & 0.349 & 0.637 & 0.269 & 0.490 & 0.914 & 0.268 \\

& BEVDet4D$^\dagger$~\cite{wang2022probabilistic} & Swin-B & 24 & 0.515 & 0.396 & \textbf{0.619} & \textbf{0.260} & 0.361 & 0.399 & 0.189 \\

\midrule

\multirow{3}*{Query-based} & DETR3D$^\dagger$~\cite{wang2022detr3d} & ResNet-101 & 24 & 0.425 & 0.346 & 0.773 & 0.268 & 0.383 & 0.842 & 0.216 \\

& DETR4D$^\dagger$~\cite{luo2022detr4d} & ResNet-101 & 24 & 0.509 & \textbf{0.422} & 0.688 & 0.269 & 0.388 & 0.496 & 0.184 \\

& PETRv2$^\dagger$~\cite{liu2022petrv2} & ResNet-101 & 24 & 0.524 & 0.421 & 0.681 & 0.267 & 0.357 & 0.377 & 0.186 \\

\midrule

\multirow{5}*{Query-BEV} & BEVFormer-S~\cite{li2022bevformer} & ResNet-101 & 24  & 0.448 & 0.375 & 0.725 & 0.272 & 0.391 & 0.802 & 0.200 \\

 & BEVFormer~\cite{li2022bevformer} & ResNet-101 & 24  & 0.517 & 0.416 & 0.673 & 0.274 & 0.372 & 0.394 & 0.198 \\

& \cellcolor[HTML]{efefef} \textbf{OCBEV}~(Ours) & \cellcolor[HTML]{efefef}ResNet-101 & \cellcolor[HTML]{efefef}12 &\cellcolor[HTML]{efefef}0.523 & \cellcolor[HTML]{efefef} 0.408 & \cellcolor[HTML]{efefef} 0.633 & \cellcolor[HTML]{efefef} 0.270 & \cellcolor[HTML]{efefef} 0.377 & \cellcolor[HTML]{efefef} \textbf{0.333} & \cellcolor[HTML]{efefef} 0.194 \\

& \cellcolor[HTML]{e2e2e2} \textbf{OCBEV}~(Ours) & \cellcolor[HTML]{e2e2e2}ResNet-101 & \cellcolor[HTML]{e2e2e2}24 & \cellcolor[HTML]{e2e2e2}\textbf{0.532} & \cellcolor[HTML]{e2e2e2}0.417 & \cellcolor[HTML]{e2e2e2}0.629 & \cellcolor[HTML]{e2e2e2}0.273 & \cellcolor[HTML]{e2e2e2}\textbf{0.339} & \cellcolor[HTML]{e2e2e2}0.342 & \cellcolor[HTML]{e2e2e2}0.187 \\

\bottomrule
\end{tabular}
    \vspace{-5pt}
    \caption{\textbf{Results comparison on the nuScenes detection validation set.} We conduct a comprehensive comparison of our OCBEV with four styles of vision-based 3D detectors. Notably, our model achieves SOTA and even outperforms latest models with data augmentation strategy CBGS~\cite{zhu2019class} as indicated by $\dagger$. Also it achieves superior performance compared to BEVFormer~\cite{li2022bevformer} NDS result even with only 12 epochs of training, and extends this performance gap with a standard 24 epochs training schedule. Meanwhile it performs best on kinestate metrics such as mAOE and mAVE, underscoring our Object-Centric approach has the capability to model moving objects in the sparse autonomous driving scene.}
    \label{t1_main_results}
\end{center}
\vspace{-15pt}
\subfloat[\textbf{Object Aligned Temporal Fusion.} We first enable ego-motion fusion and then object-motion fusion by increasing the number of aligned objects. Results indicate that Object Aligned Temporal Fusion works effectively with ego-motion fusion and achieves the best performance with 30 aligned objects.
\vspace{-2mm}
\label{tab:ablation1}
]
{
\begin{minipage}{\linewidth}{\begin{center}
\tablestyle{10pt}{1.02}
\begin{tabular}{l |c c|c c| c c c c c }
\toprule
Backbone & \textbf{Global} & \textbf{Num. Objects} & NDS~$\uparrow$  & mAP~$\uparrow$  & mATE~$\downarrow$     & mASE~$\downarrow$     & mAOE~$\downarrow$     & mAVE~$\downarrow$    & mAAE~$\downarrow$    \\
\midrule

ResNet-101 &  & - & 0.371 & 0.316 & 0.798 & 0.292 & 0.605 & 0.919 & 0.252 \\

ResNet-101 & \checkmark & - & 0.401 & 0.308 & 0.801 & 0.288 & 0.555 & 0.669 & 0.231 \\

ResNet-101 & \checkmark & 10 & 0.403 & 0.303 & 0.826 & 0.289 & 0.533 & 0.622 & 0.220 \\

\rowcolor{bar!10}
ResNet-101 & \checkmark & 30 & \textbf{0.413} & \textbf{0.323} & 0.801 & 0.285 & 0.533 & 0.640 & 0.232 \\

ResNet-101 & \checkmark & 50 & 0.409 & 0.318 & 0.809 & 0.293 & 0.524 & 0.647 & 0.230 \\
\bottomrule
\end{tabular}
\end{center}}
\end{minipage}
}\\
\centering
\subfloat[\label{tab:ablation2}\textbf{Object Focused Multi-View Sampling.} Our proposed model benefits significantly from the multi-view sampling strategy, particularly within the usual object height range of -2 meters to 2 meters. Moreover adaptive local height takes performance a step further.  
]
{
\centering
\begin{minipage}{0.42\linewidth}{\begin{center}
\tablestyle{10pt}{1.02}

\begin{tabular}{l |c|c c}
\toprule
Backbone  & \textbf{Height Range} & NDS~$\uparrow$  & mAP~$\uparrow$ \\
\midrule
ResNet-101 & - & 0.371 & 0.316  \\
ResNet-101 & [-5m, -2m] & 0.412 & 0.321  \\
ResNet-101 & [-2m, ~2m] & 0.421 & 0.325 \\
ResNet-101 & [~2m, ~5m] & 0.410 & 0.325 \\
\rowcolor{bar!10}
ResNet-101 & adaptive & \textbf{0.430} & \textbf{0.324} \\
\bottomrule
\end{tabular}

\end{center}}\end{minipage}
}
\hspace{8mm}
\subfloat[\textbf{Object Informed Query Enhancement.} Our model exhibits improved performance with the additional heatmap supervision. Further, the replacement of predefined queries of points with object positional information results in a further boost in performance until the replacement points number reaches 50.
\label{tab:ablation3}
]
{
\begin{minipage}{0.5\linewidth}{\begin{center}
\tablestyle{9pt}{1.02}
\begin{tabular}{l |c c|c c }
\toprule
Backbone & \textbf{Heatmap} & \textbf{Num. Points} & NDS~$\uparrow$  & mAP~$\uparrow$ \\
\midrule
ResNet-101 &  & - & 0.371 & 0.316 \\
ResNet-101 & \checkmark & - & 0.414 & 0.292 \\
ResNet-101 & \checkmark & 30 & 0.412 & 0.327 \\
\rowcolor{bar!10}
ResNet-101 & \checkmark & 50 & \textbf{0.422} & \textbf{0.338} \\
ResNet-101 & \checkmark & 70 & 0.416 & 0.320 \\
\bottomrule
\end{tabular}
\end{center}}\end{minipage}
}
\vspace{-5pt}
\caption{
\textbf{Ablation study.} We conducted ablation experiments on a lighting version OFBEV model trained with a 12 epochs which remove temporal attention and reduce the spatial attention layer from 6 to 3. Our results indicate that each object-centric design has a positive impact on the overall performance of the model. More results are in the supplementary material.
}
\label{tab:ablations} 
\vspace{-15pt}
\end{table*}

\subsection{Implementation Details}
\noindent\textbf{Dataset and metrics.} To demonstrate the effectiveness of the proposed method, we conduct experimental evaluations on the challenging nuScenes 3D detection benchmark~\cite{caesar2020nuscenes}. The dataset consists of 1000 scenes, each of which is about 20 seconds, and the sample rate is 2 HZ (40 annotation samples/scene). Each sample contains six images from six 360-degree cameras. It is split into 700/150/150 videos (28k/6k/6k samples) for training, validation, and testing, respectively. Ten categories, including car, bus, and pedestrian are adopted for evaluation. For the 3D object detection metrics, mean average precision (mAP) is computed over four thresholds using center distance on the ground plane. There are five other types of true positive metrics (TP metrics), including mean Average Translation Error (mATE), mean Average Scale Error (mASE), mean Average Orientation Error(mAOE), mean Average Velocity Error(mAVE), and mean Average Attribute Error(mAAE). Furthermore, the nuScenes detection score (NDS) is a comprehensive and the most important metric that combines all aspects above.

\mypara{Experimental settings.} 
The input image size is $ 1600 \times 900$ as default. Following previous practice~\cite{li2022bevformer, wang2022detr3d, luo2022detr4d}, our model adopts the FCOS3D~\cite{wang2021fcos3d} pre-trained ResNet-101~\cite{he2016deep} as backbone and initialized weights. After the backbone, we use a FPN~\cite{lin2017feature} structure as the neck with the dimension of 256 and the size of \nicefrac{1}{16}, \nicefrac{1}{32}, \nicefrac{1}{64}. As for the transformer layer, we utilize six layers for temporal, spatial, and decoder attention. For the BEV setting, we set the size of BEV queries as 300 $\times$ 300 (A minor variable, 200: $53.1\%$ NDS vs. 300: $53.2\%$ NDS ), the perception range as $[-51.2m, 51.2m]$ for the $XY$ plane, and the height range as $[-5m, 3m]$. For the training setting, we train our models for 12 and 24 epochs. The initial learning rate is set as $2 \times 10^{-4}$ and a cosine learning rate decay policy is adopted.

\subsection{Main Results}
We compare our experimental results with other vision-based 3D detection methods on the nuScenes validation set, especially those query-based vision methods. The results are shown in Table~\ref{t1_main_results}. Compared to the current \textit{state-of-the-art} query-based method BEVFormer~\cite{li2022bevformer}, OCBEV outperforms it over $1.5$ points on the validation set. Moreover, OCBEV outperforms DETR4D~\cite{luo2022detr4d} over 2.3 points and PETRv2~\cite{liu2022petrv2} over 0.8 points with the same backbone and training schedule. For the depth-based methods, OCBEV is 1.7 points higher than the BEVDet4d~\cite{huang2022bevdet4d}. All these results prove that focusing on objects can bring huge improvement, and our proposed OCBEV is a strong 3D outdoor detector and push the
previous SOTA to 53.2\%.

\begin{figure}[t!]
    \centering
    \includegraphics[width=0.98\linewidth]{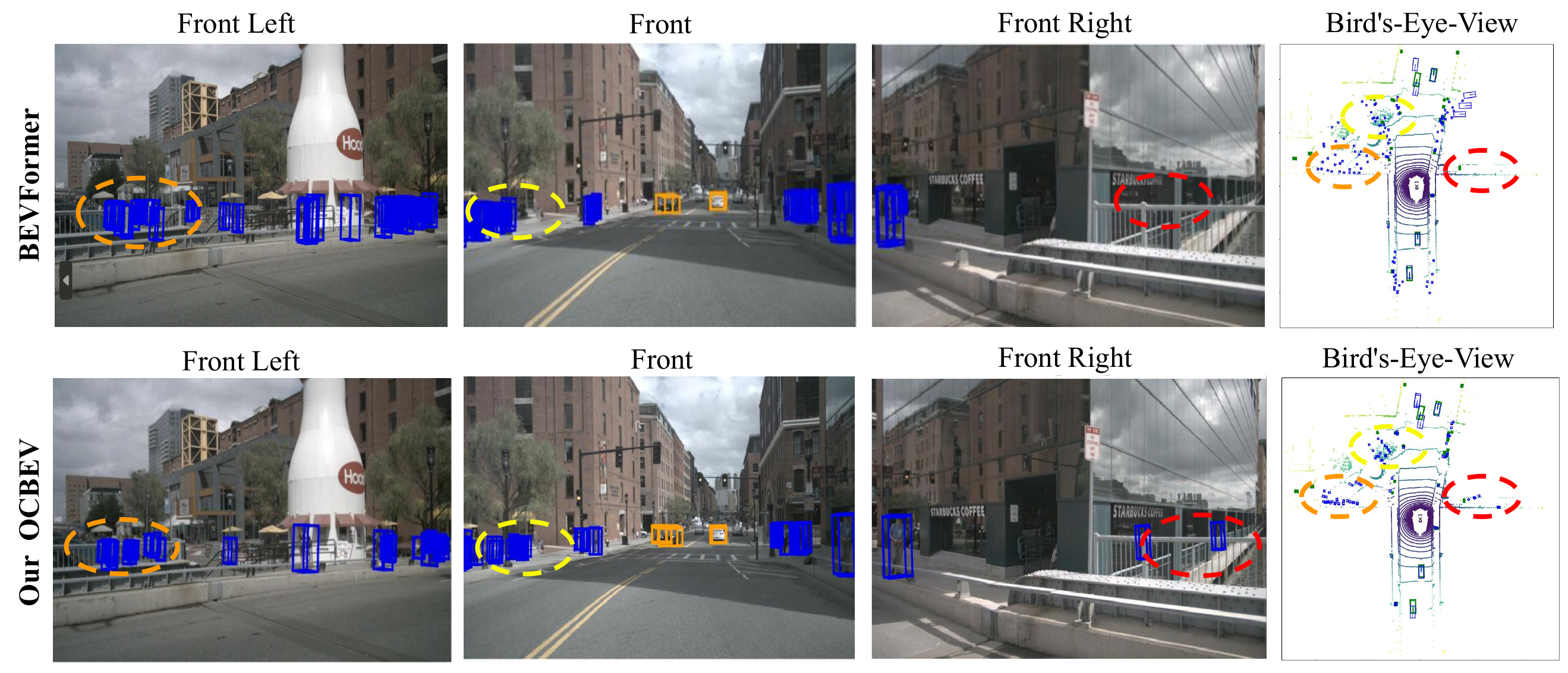}
    \vspace{-5pt}
    \caption{\textbf{Qualitive Results compared with BEVFormer.} Our model predicts more accurately than BEVFormer which successfully detects the pedestrian in the front right view and less false positive cases in the front right view.}
    \label{fig:f5_visualization}
    \vspace{-10pt}
\end{figure}

Moreover, OCBEV can converge faster than existing methods. With only half of the training schedule (12 epochs), the NDS of OCBEV is $0.6$ points higher than BEVFormer~\cite{li2022bevformer} with 24 epochs ($52.3\%$ NDS vs. $51.7\%$ NDS). And our 12-epoch result has already gone beyond most methods, which means the encoder positional hint to the decoder can help the network to converge faster largely. The benefits of the proposed Object-Centric modules are also revealed in the kinematic metrics. Metrics relying on the kinestate such as orientation (mAOE: $33.9\%$) and velocity (mAVE: $33.3\%$ and $34.2\%$) outperform BEVFormer~\cite{li2022bevformer} and get best results among all existing methods. It also gets better results on the translation (mATE: $62.9\%$). This demonstrates that our method pays more attention to moving objects and can naturally improve overall performance. Some qualitative results are shown in Fig.~\ref{fig:f5_visualization} and show that our model can give more precise 3D bounding boxes.

\vspace{-5pt}
\subsection{Ablation Studies}
\vspace{-5pt}
To verify the effectiveness of the proposed three Object-Centric modules, we conduct ablation experiments on a lighting version OFBEV model trained with a 12 epochs training schedule and reduce the attention layer from 6 to 3 to save the memory. Results are reported in Table~\ref{tab:ablations}.

\mypara{Object Aligned Temporal Fusion.}
To study the effect of the Object Aligned Temporal Fusion module, we remove the ego-motion and object ego-motion aligned fusion. Note that we also remove the temporal attention between the current and previous BEV feature. Then, we add ego-motion and different numbers of objects for object ego-motion aligned fusion step by step. As shown in Table~\ref{tab:ablation1}, by adding the global alignment, the NDS increase by $0.5\%$ points. Regarding the object ego-motion aligned fusion module, the gain increases from modeling 10 objects to 30 objects and drops from 30 to 50 objects. We owe this phenomenon to that one scene may not have so many objects. Thus in this setting, modeling 30 objects can get the best results and outperform by $1.7\%$ points than that without object-focused alignment fusion.

\begin{table}[t!]
    \centering
    \vspace{-5pt}
    \tablestyle{6pt}{1.02}
    \begin{tabular}{l|c c c|c c}
\toprule
Backbone & \textbf{Temporal} & \textbf{Spacial} & \textbf{Decoder} & NDS$\uparrow$  & mAP$\uparrow$    \\
\midrule

ResNet-101 &  &  &  & 0.371 & 0.316 \\

\midrule

ResNet-101 & \checkmark &  &  & 0.413 & 0.323 \\

ResNet-101 &  & \checkmark &  & 0.430 & 0.324 \\

ResNet-101 &  &  & \checkmark & 0.422 & 0.338 \\

\midrule

ResNet-101 & \checkmark & \checkmark &  & 0.439 & 0.323 \\

ResNet-101 & \checkmark &  & \checkmark & 0.439 & 0.339 \\

ResNet-101 &  & \checkmark & \checkmark & 0.444 & 0.335 \\

\midrule

ResNet-101 & \checkmark & \checkmark & \checkmark & \textbf{0.448} & \textbf{0.333} \\

\bottomrule
\end{tabular}
    \caption{\textbf{Combination of object-centric components.} Temporal, Spatial and Decoder represents \textit{Object Aligned Temporal Fusion}, \textit{Object Focused Multi-View Sampling} and \textit{Object Informed Query Enhancement} respectively. Performance increases with the number of modules increase. } 
    \label{t3_combination}
    \vspace{-5mm}
\end{table}

\mypara{Object Focused Multi-View Sampling.}
In the nuScenes validation set, we use the point cloud range of $[-5m, 3m]$ as the global height. In the spatial attention, apart from sample points in the global height range. We also sample points in the following local height ranges $[-5m, -2m]$, $[-2m, 2m]$ and $[2m, 3m]$. These three ranges are the bottom, middle, and top of the 3D scene, respectively. Also, we study the effect of making the height adaptive. The results are shown in Table~\ref{tab:ablation2}, which indicate that paying attention to object-dense local height $[-2m, 2m]$ (NDS: +$5.0\%$) can have the most gain compared to other less-object-dense height ranges. While the adaptive height range will gain the most (NDS: +$5.9\%$) and be much higher than other ranges.

\begin{figure}[t]
    \centering
    \includegraphics[width=0.95\linewidth]{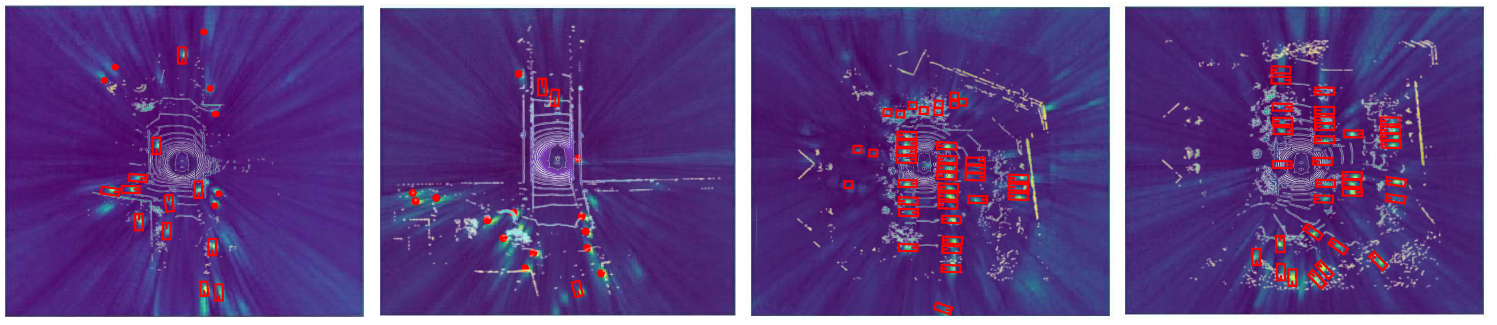}
    \vspace{-5pt}
    \caption{\textbf{Visualization of the heatmap.} It shows heatmaps of two scenes with ground truth boxes where red points are pedestrians and bounding boxes are cars. Bright spots (high confidence locations) correspond to the absolute position of objects, which gives the decoder a precise positional hint.}
    \label{fig:f6_heatmap}
    \vspace{-15pt}
\end{figure}

\mypara{Object Informed Query Enhancement.}
In order to validate the effectiveness of object-centric encoder output for the decoder, we remove the heatmap supervision and all encoder positional hints as the baseline. We then add the supervision and replace the predefined positional embeddings step by step. Results in Table~\ref{tab:ablation3} show that by adding heatmap supervision, the NDS increases by 4.3 points and achieves $41.4\%$. Moreover, replacing pre-defined reference points and positional embeddings will gain further improvements and reach a peak with a replacement points number of 50 and finally lead to a result of 42.2\% NDS. Fig.~\ref{fig:f6_heatmap} shows the visualization of our heatmaps from four different scenes. The high confidence position shown as bright spots corresponds to the absolute position of objects, which gives the decoder a precise positional hint. All prove the effectiveness of our Object Informed Query Enhancement.

\mypara{Combination of object-centric components.}
In Table~\ref{t3_combination}, we ablate the combination of three object-centric modules to confirm their contributions to the final results. The baseline is the same as mentioned before, we remove all our modules and the temporal attention. \textit{Temp} means the Object Aligned Temporal Fusion, \textit{Spatial} denotes Object-Focused Temporal Alignment, and \textit{Decoder} represents Object Informed Query Enhancement for the decoder. Obviously, the performance will increase along with the number of modules increase. Finally, the complete version with all three modules outperforms 7.7 points compared to the baseline and reaches 44.8\% NDS with only 12 epochs of training.

\section{Conclusion}
\label{sec:conclusion}
Query-based BEV paradigm benefits from both BEV’s perception power and end to-end pipeline. Most 3D BEV detectors ignore moving objects in the scene and leverage temporal and spatial information globally. In this work, we propose a novel Object-Centric query-based BEV detector OCBEV. Three modules: Object Aligned Temporal Fusion, Object Focused Multi-View Sampling and Object Informed Query Enhancement focus on instance-level temporal modeling, spatial exploition and decoder respectively. Our results achieve a state-of-the-art result on nuScenes benchmark and have a faster convergence speed as it only needs half training iterations to get comparable performance.

\newpage
\setcounter{table}{0}
\setcounter{figure}{0}
\renewcommand{\thetable}{S\arabic{table}}
\renewcommand\thefigure{S\arabic{figure}}

\vspace{12pt}
\twocolumn[{
\renewcommand\twocolumn[1][]{#1}
\maketitle
\begin{center}
    \centering
    {\Large \textbf{OCBEV: Object-Centric BEV Transformer for Multi-View 3D Object Detection 
    \vspace{15pt}
    \\ (Supplementary Material)}}
    \vspace{20pt}
    \begin{center}
    \hrule\hrule\hrule
    \hypersetup{linkcolor=black}
    \vspace{20pt}
    \textbf{\Large Content} \\
    \vspace{10pt}
    \textbf{\large A. Additional Related Work} \dotfill \pageref{sec:sup_A_add_related} \\
    \vspace{7pt}
        \hspace{10pt} A.1 Three Types of Multi-view 3D Object Detection \dotfill \pageref{sec:sup_A1_3_types_detection} \\
        \vspace{7pt}
        \hspace{10pt} A.2 One-stage and Two-stage DETR Style Decoder \dotfill \pageref{sec:sup_A2_3_DETR_Decoder} \\
        \vspace{7pt}
    \textbf{\large B. Implementation Details} \dotfill \pageref{sec:sup_B_implementation} \\
    \vspace{7pt}
        \hspace{10pt} B.1 Glossary of Notation \dotfill \pageref{sec:sup_B1_types_Glossary of Notation} \\
        \vspace{7pt}
        \hspace{10pt} B.2 Network Architecture \dotfill \pageref{sec:sup_B2_Network Architecture} \\
        \vspace{7pt}
        \hspace{10pt} B.3 Training Setting \dotfill \pageref{sec:sup_B3_Training Setting} \\
        \vspace{7pt}
        \hspace{10pt} B.4 Dataset Pre-processing \dotfill \pageref{sec:sup_B4_Dataset Pre-processing} \\
        \vspace{7pt}
    
    \textbf{\large C. Extended Experiment Results} \dotfill \pageref{sec:sup_C_extended_experiment} \\
    \vspace{5pt}
        \hspace{10pt} C.1. Analysis of Categories in Main Results \dotfill \pageref{sec:sup_C1_Analysis of Categories in Main Results} \\
        \vspace{7pt}
        \hspace{10pt} C.2. Additional Ablation Studies \dotfill \pageref{sec:sup_C2_Additional Ablation Studies} \\
        \vspace{7pt}
    
    \textbf{\large D. Visualization} \dotfill \pageref{sec:sup_D_visualization} \\
    \vspace{7pt}

    \vspace{12pt}
    \hrule\hrule\hrule
    \hypersetup{linkcolor=red}
    \end{center}
    \label{fig:teaser}
\end{center}
}]

\appendix
\section{Additional Related Work}
\label{sec:sup_A_add_related}
\subsection{Three Types of Multi-view 3D Object Detection}
\label{sec:sup_A1_3_types_detection}
\begin{figure}[tbp]
\centering
\subfloat[Object-query-based]{\label{fig:sup_f1a_classification}\includegraphics[width=0.31\linewidth]{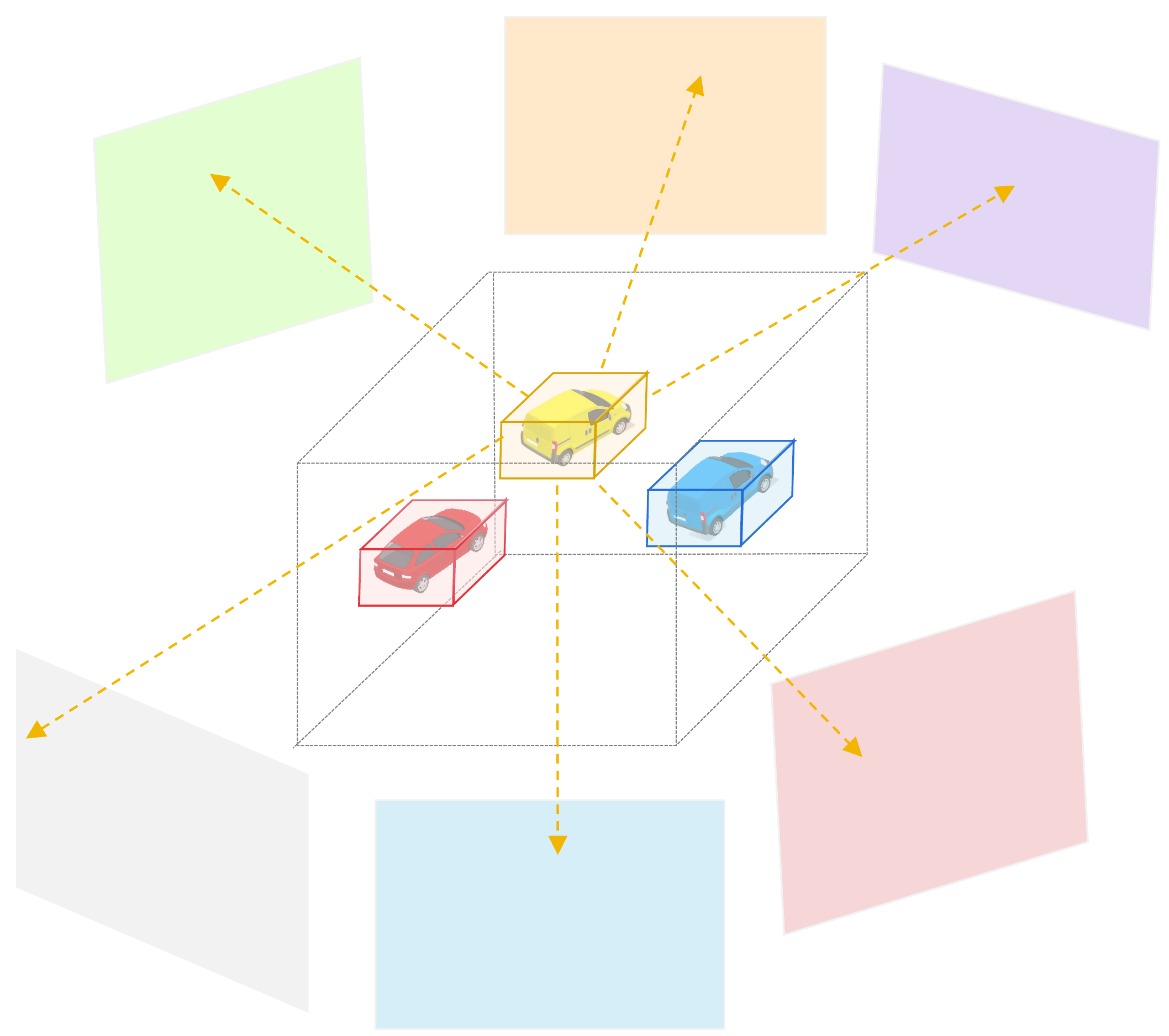}}
\subfloat[BEV-query-based]{\label{fig:sup_f1b_classification}\includegraphics[width=0.31\linewidth]{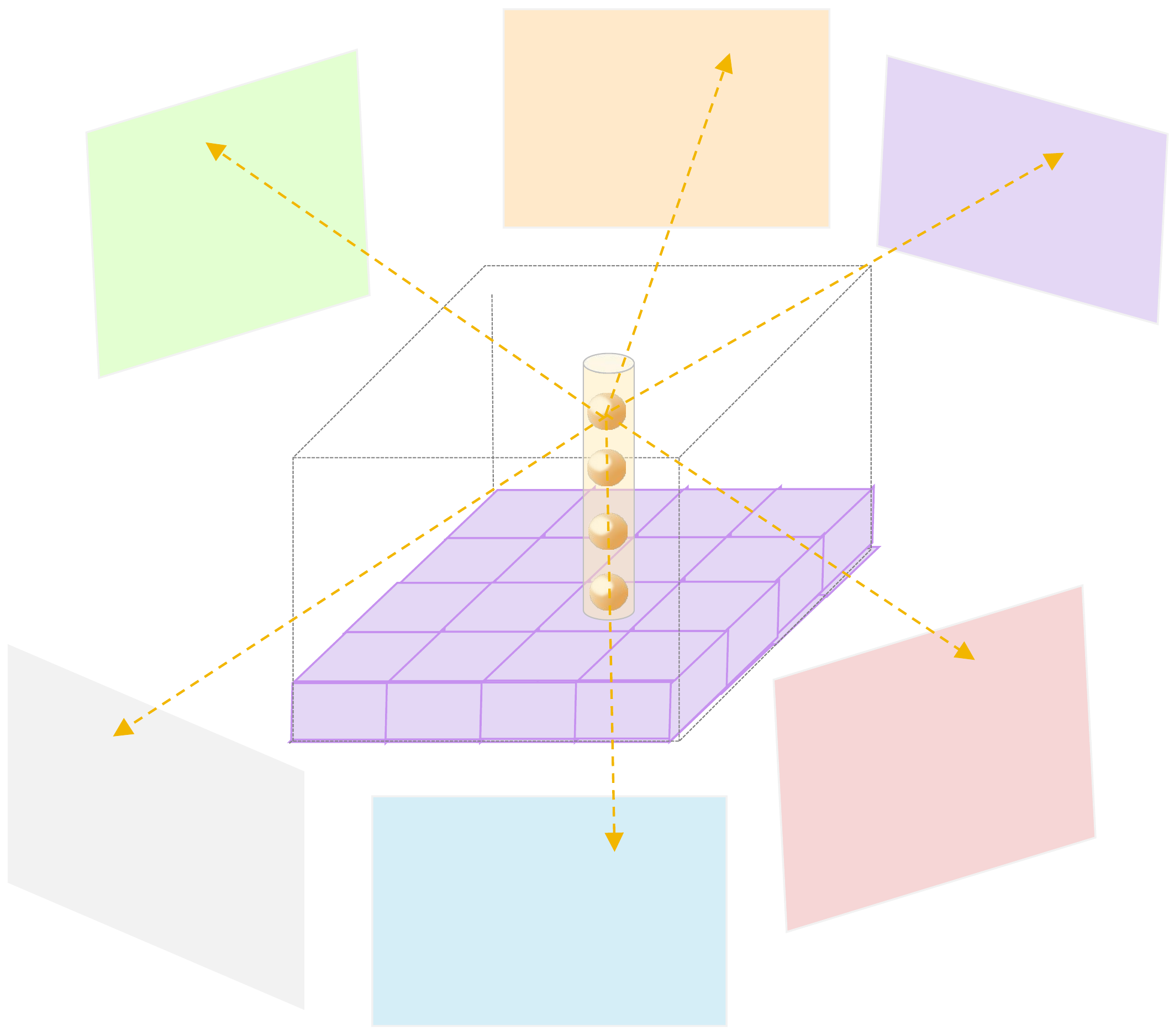}}
\subfloat[BEV-depth-based]{\label{fig:sup_f1c_classification}\includegraphics[width=0.31\linewidth]{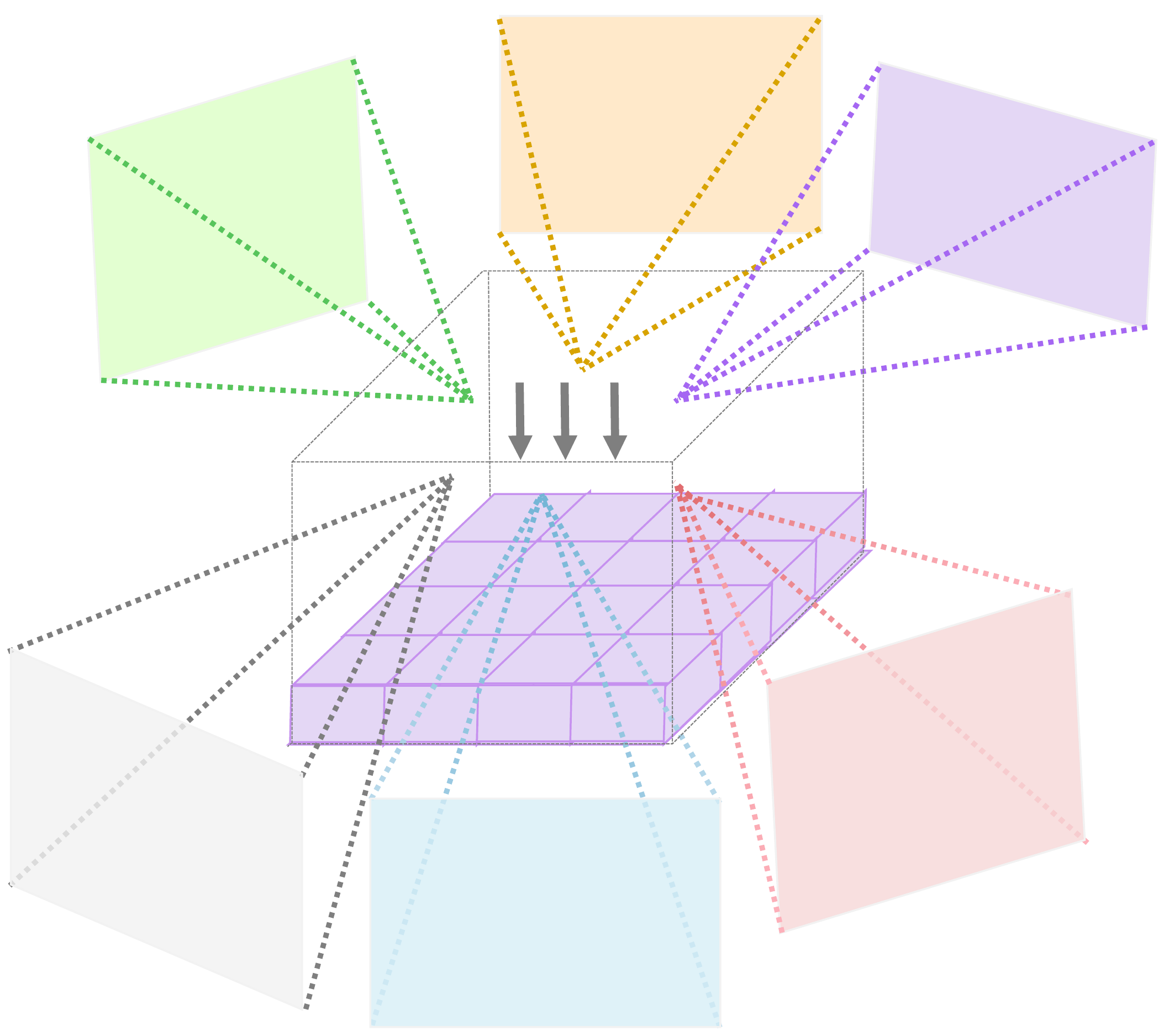}}\\
\vspace{-2mm}
\caption{Three types of Multi-view 3D Detectors. The (a) Object-query-based and (b) BEV-query-based together form the query-based methods. The (b) BEV-query-based and (c) BEV-depth-based form the BEV-based methods.}
\label{fig:sup_f1}
\vspace{-6mm}
\end{figure}

Here we will give a more detailed explanation on the classification of multi-view vision-based detectors. We first divided the multi-view vision-based detector into object-(query)-based methods~\cite{wang2022detr3d} and BEV-based methods, and the classification is based on the format of features. Object-based approaches also called object-query-based methods use an object in 3D space as a query, and obtain temporal and spatial information by querying historical queries and 2D images as shown in Fig.~\ref{fig:sup_f1}. We can know that this type of method focuses on the object. However, due to the sparsity and scattered distribution of objects in the scene, accomplishing a meticulous one-to-one correlation between the physical entities and the related queries can be a daunting and demanding task. So BEV-based methods are proposed as BEV has strong representation capability. 

Similar to the object-query-based methods, we also need to consider BEV to obtain temporal and spatial information. According to the way to extract information from perspective-view image features. Multi-view vision-based detectors are divided into query-based methods and (BEV-)depth-based methods. BEV-depth-based methods~\cite{huang2021bevdet} predict a latent depth to build the multi-view frusta and splat them to form the BEV while both object-query-based methods and BEV-query-based methods~\cite{li2022bevformer} pre-define learnable queries and using 3D points projecting to 2D features. In conclusion, there are three types of methods: \textit{\textbf{object-query-based}}, \textit{\textbf{BEV-query-based}}, and \textit{\textbf{BEV-depth-based methods}}. The former two comprise query-based methods, and the latter two are BEV-based methods.

\begin{figure}[tbp]
\centering
\subfloat[One-stage Decoder]{\label{sup_f2a_one_stage}\includegraphics[width=0.35\linewidth]{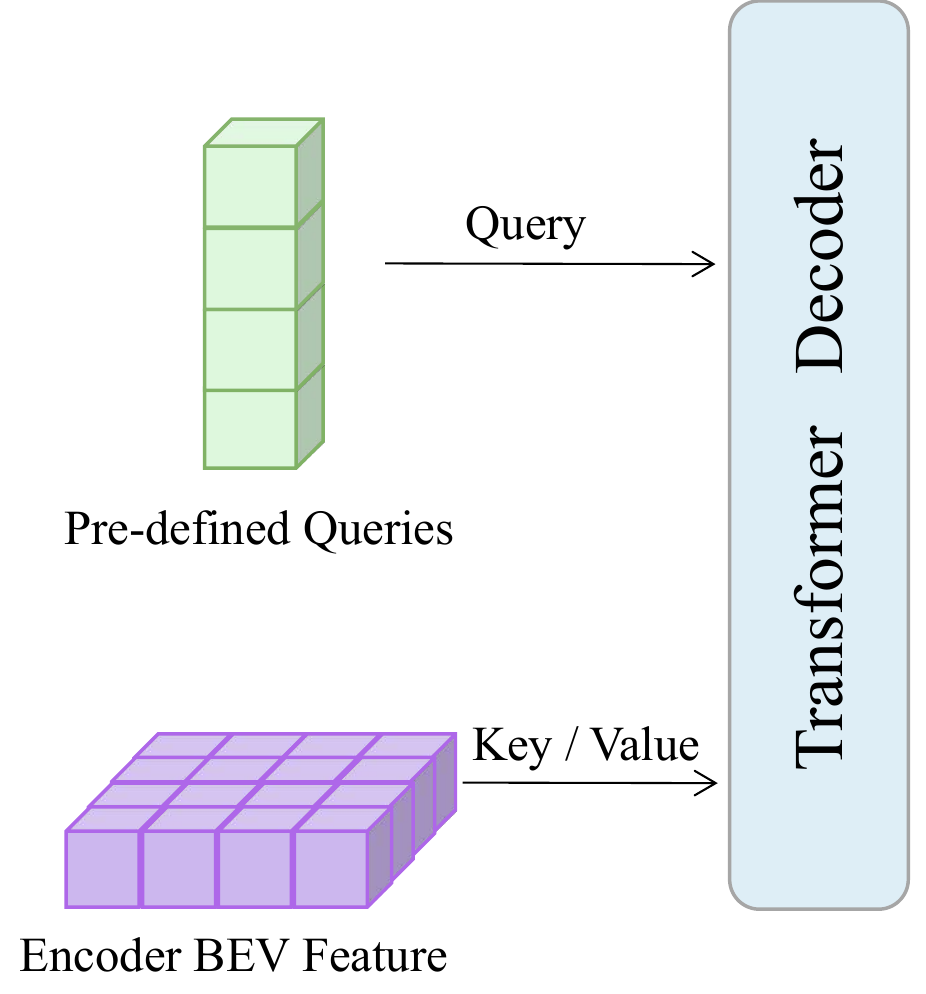}}
\hspace{6mm}
\subfloat[Two-stage Decoder]{\label{sup_f2b_two_stage}\includegraphics[width=0.42\linewidth]{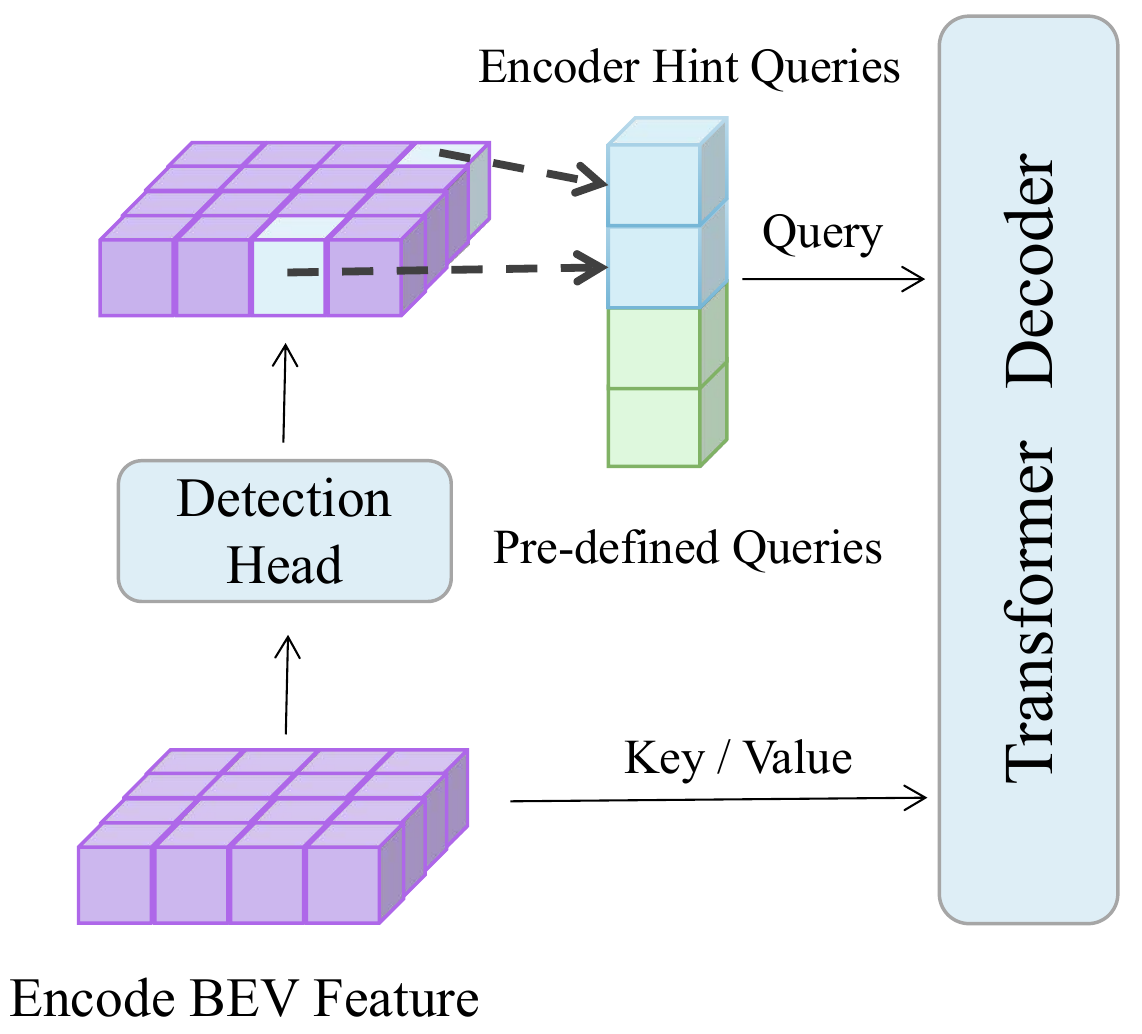}}

\caption{\textbf{One-stage and two-stage DETR style decoder.} (a) One-stage decoder uses the pre-defined queries to match the encoder output while (b) two-stage decoder use mixed queries while encoder hint queries are provided by the detection results after the encoder. These high-confidence locations are similar to proposals in two-stage CNN methods.}
\label{fig:sup_f2_decoder}
\vspace{-2mm}
\end{figure}

\subsection{One-stage and Two-stage DETR Style Decoder}
\label{sec:sup_A2_3_DETR_Decoder}
The original one-stage and two-stage DETR~\cite{carion2020end} style decoders are shown in Fig.~\ref{fig:sup_f2_decoder}. For the DETR~\cite{carion2020end} style decoder, the common one-stage approach~\cite{carion2020end} is to pre-define a certain number of queries as Fig.~\ref{sup_f2a_one_stage}. These queries correspond to the features of the encoder output. But it is difficult to find a direct match between the predefined queries and encoder output. Thus, inspired by the traditional CNN-based two-stage target detection, we can give rough results before giving final predictions. Two-stage DETR~\cite{carion2020end} style decoder~\cite{yang2022bevformer} is such that, as shown in Fig.~\ref{sup_f2b_two_stage}, after encoder, we first use the detection head to make preliminary predictions on the encoder output and feed the detection high score results as queries of the decoder into the decoder.

\section{Implementation Details}
\label{sec:sup_B_implementation}
\subsection{Glossary of Notation}
\label{sec:sup_B1_types_Glossary of Notation}
\begin{table*}
    \centering
    \vspace{-10pt}
    \tablestyle{3pt}{0.5}
    \begin{tabular}{c|l|c}
\toprule
\textbf{Notation} & \multicolumn{1}{|c|}{\bfseries Meaning\mdseries} & \textbf{Shape} \\
\midrule

$\mQ_t$ & BEV queries at timestamp $t$. & $C_{q}\!\times\!H_{q}\!\times\!W_{q}$ \\
\midrule

$\mQ_{t^\prime}$ & BEV queries at timestamp $t-1$. & $C_{q}\!\times\!H_{q}\!\times\!W_{q}$ \\
\midrule

$\overline{\mQ}_{t^\prime}$ & Flatten version of BEV queries at timestamp $t$. & $C_{q}\!\times\!(H_{q}W_{q})$ \\
\midrule

$\overline{\mQ}_{t^\prime}$ & Flatten version of BEV queries at timestamp $t-1$. & $C_{q}\!\times\!(H_{q}W_{q})$ \\
\midrule\midrule

$\sP$ & Ego-motion overlapping part matching relationship between timestamp $t-1$ and $t$. & $N_p$ pairs\\
\midrule

$\mI$ & Overlapping part matching indices at timestamp $t-1$. & $1\!\times\!N_p$ \\
\midrule

$\mJ$ & Overlapping part matching indices at timestamp $t$. & $1\!\times\!N_p$ \\
\midrule

$\mathcal{R}_{t^\prime \rightarrow t}$ & Rotation matrix from timestamp $t-1$ to $t$. & $3\!\times\!3$ \\
\midrule

$\mathcal{R}_{t \rightarrow t^\prime}$ & Rotation matrix from timestamp $t$ to $t-1$. & $3\!\times\!3$ \\
\midrule
 
$\gamma_{t^\prime \rightarrow t}$ & Yaw angle of rotation matrix from timestamp $t-1$ to $t$. & $\mathcal{R}$ \\
\midrule

$\gamma_{t \rightarrow t^\prime}$ & Yaw angle of rotation matrix from timestamp $t$ to $t-1$. & $\mathcal{R}$ \\
\midrule

$\mathcal{T}_{t^\prime \rightarrow t}$ & Translation vector from timestamp $t-1$ to $t$ & $3\!\times\!1$ \\
\midrule

$\mathcal{T}_{t \rightarrow t^\prime}$ & Translation vector from timestamp $t$ to $t-1$. & $3\!\times\!1$ \\
\midrule

$\vg_r$ & \textbf{Function}: Rotation operation. & - \\
\midrule

$\vg_t$ & \textbf{Function}: Translation operation. & - \\
\midrule

$\vf_{a}^{ego}$ & \textbf{Function}: Ego-motion temporal alignment and fusion. & - \\
\midrule

$\mQ_{a}^{ego}$ & Aligned and fused BEV queries by Ego-motion temporal alignment and fusion. & $C_{q}\!\times\!H_{q}\!\times\!W_{q}$ \\
\midrule\midrule

$\vp_{t^\prime}^{m}$ & $xy$ coordinates of moving objects in timestamp $t-1$ and $\mQ_{t^\prime}$ reference frame. & $N_m\!\times\!2$ \\
\midrule

$\mI_{t^\prime}^{m}$ & $\mQ_{t^\prime}$ indices of moving objects in timestamp $t-1$ and $\mQ_{t^\prime}$ reference frame. & $1\!\times\!N_m$ \\
\midrule

$\vv_{t^\prime}^{m}$ & $xy$ velocity of moving objects in timestamp $t-1$. & $N_m\!\times\!2$ \\
\midrule

$\vp_{t}^{m}$ & $xy$ coordinates of moving objects in timestamp $t$ and $\mQ_{t^\prime}$ reference frame. & $N_m\!\times\!2$ \\
\midrule

$\mI_{t}^{m}$ & $\mQ_{t^\prime}$ indices of moving objects in timestamp $t$ and $\mQ_{t^\prime}$ reference frame. & $1\!\times\!N_m$ \\
\midrule

$\mJ_{t}^{m}$ & $\mQ_{t}$ indices of moving objects in timestamp $t$ and $\mQ_{t}$ reference frame. & $1\!\times\!N_m$ \\
\midrule

$\vf_{a}^{obj}$ & \textbf{Function}: Object-motion temporal alignment and fusion. & - \\
\midrule

$\mQ_{a}^{obj}$ & Aligned and fused BEV queries by Object-motion temporal alignment and fusion. & $C_{q}\!\times\!H_{q}\!\times\!W_{q}$ \\
\midrule

$\mQ_{a}$ & Aligned and fused BEV queries by Object Aligned Temporal Fusion. & $C_{q}\!\times\!H_{q}\!\times\!W_{q}$ \\
\midrule\midrule

DAttn & \textbf{Function}: Deformable attention. & - \\
\midrule

SpaA & \textbf{Function}: Spatial Attention between BEV and image features. & - \\
\midrule

$\mF_{t}$ & Image features at timestamp $t$. & $C_{f}\!\times\!H_{f}\!\times\!W_{f}$ \\
\midrule

$\textbf{Z}_h$ & Height range for the 3D reference points & $[\mathcal{R}_1, \mathcal{R}_2]$ \\
\midrule

$\vp_{3d}$ & 3D reference points for Spatial Attention. & $N_h\!\times\!3$ \\
\midrule

$g_p$ & \textbf{Function}: Projection 3D points to 2D image feature. & - \\
\midrule

$\textbf{Z}_{h,g}$ & Global height range for reference points & $[\mathcal{R}_1, \mathcal{R}_2]$ \\
\midrule

$\textbf{Z}_{h,l}$ & Local height range for reference points & $[\mathcal{R}_1, \mathcal{R}_2]$ \\
\midrule

$\hat{\textbf{Z}}_{h,l}$ & Adaptive local height range for reference points & $[\mathcal{R}_1, \mathcal{R}_2]$ \\
\midrule

OFSpaA & \textbf{Function}: Spatial Attention by Object Focused Multi-View Sampling
 & - \\
\bottomrule

\end{tabular}
    \caption{Notation table: all symbols and functions. We show the shapes for symbols. Four parts are \textit{\textbf{BEV query definition}}, \textit{\textbf{Ego-motion temporal fusion}}, \textit{\textbf{Object-motion temporal fusion}} and \textit{\textbf{Object Focused Multi-View Sampling}}, respectively.} 
    \label{tab:sup_t1_notable_table}
    \vspace{-5mm}
\end{table*}

All notations, including symbols and functions, are listed in Table~\ref{tab:sup_t1_notable_table} and in the order of appearance of the article. We explain all their definitions, and for symbols, we give their shapes while giving the inputs of all functions. The first part is the prerequisites to define the BEV queries. The two middle parts are the symbols that appear in Object Aligned Temporal Fusion. The second part is the notation used in the Ego-motion temporal fusion, and the third is for Object-motion fusion. The final part is the definition of Object Focused Multi-View Sampling and spatial exploitation.

\subsection{Network Architecture}
\label{sec:sup_B2_Network Architecture}
\noindent\textbf{Backbone.} OCBEV's image backbone ResNet-101~\cite{he2016deep} yields 3-level image feature maps of strides 16, 32, and 64. We replace the original convolution network with deformable convolutional networks v2~\cite{dai2017deformable} for only the last two layers and the deformable groups equal to 1. The FPN~\cite{lin2017feature} neck with input sizes 512, 1024, 2048 after the backbone produces 4-level features of stride 16, 32, 64, and 128. 
\begin{figure}[t]
    \centering
    \includegraphics[width=0.82\linewidth]{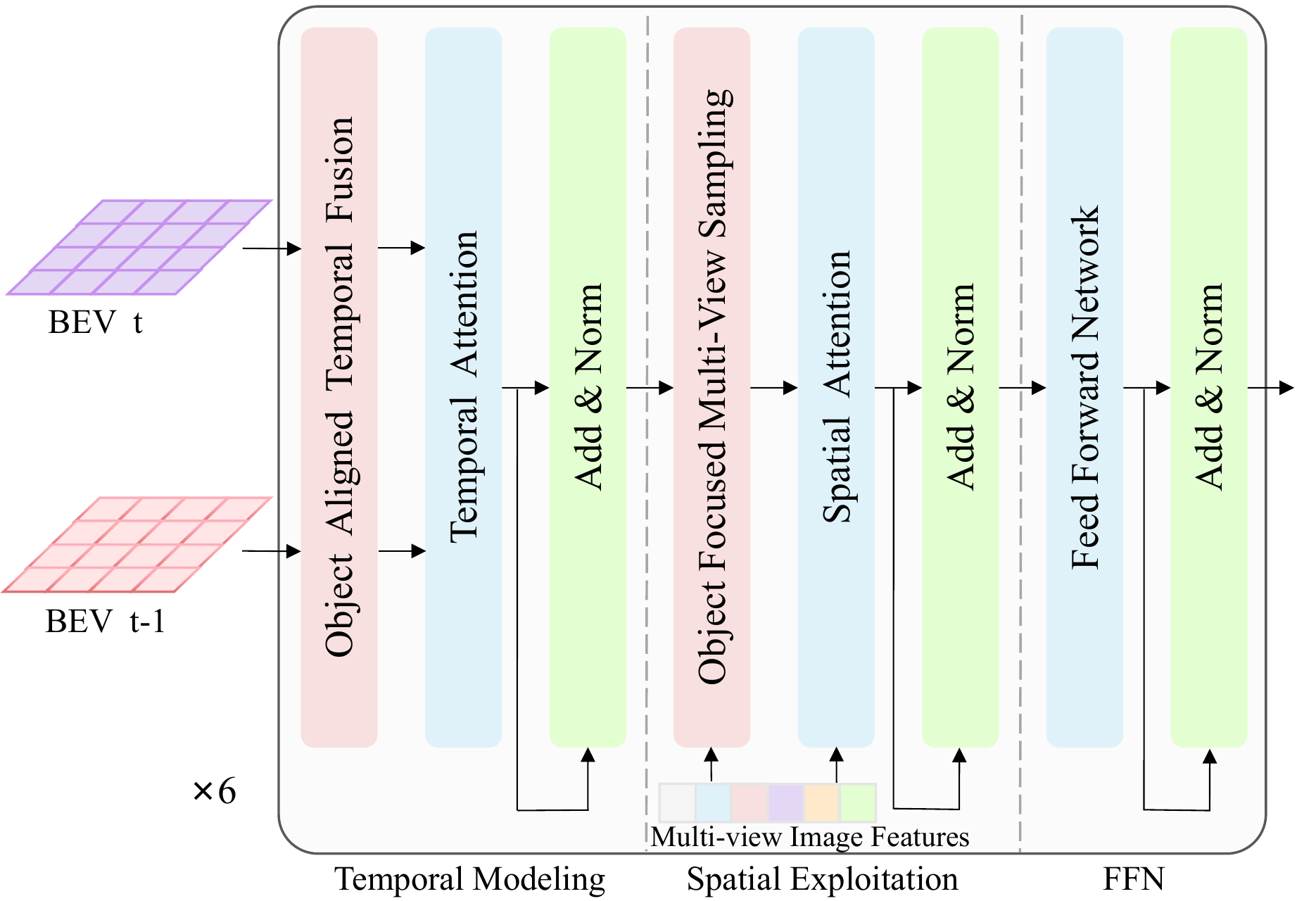}
    \vspace{-6pt}
        \caption{\textbf{The architecture of the encoder.} Red modules are added by ourselves. Blue modules are deformable attention, and green modules are \textit{Add and Norm} modules. First and middle layers are temporal modeling and spatial exploitation, respectively. Final layers are Forward Network.}
    \label{fig:sup_f3_encoder}
    \vspace{-10pt}
\end{figure}

\noindent\textbf{Encoder layers.} The architecture of the encoder is shown in Fig.~\ref{fig:sup_f3_encoder}. The dimension of BEV queries is 256. We use six layers of deformable attention~\cite{zhu2020deformable} for the encoder to do temporal modeling and spatial exploitation. Note that we only consider the range with both length and width from $-51.2m$ to $51.2m$ and the height from $-5m$ to $3m$. As shown in Fig.~\ref{fig:sup_f3_encoder}, the encoder sequence is similar to BEVFormer~\cite{li2022bevformer}. For each sub-layer, the initial three sub-layers constitute temporal modeling. Before formal temporal attention, we add our Object Aligned Temporal Fusion to align and fuse the history information. The middle three sub-layers constitute spatial exploitation. We use Object Focused Multi-View Sampling to get object-dense 3D reference points and then feed these reference points to the spatial attention. For each height range, we sample 4 points along the $z$-axis evenly. All transformer layers have eight heads. A 2-layer Feed Forward Network with a 1024-dimension hidden layer is finally implemented after them.

\noindent\textbf{Decoder layers.} Decoder consists of 6 layers, similar to the encoder. The queries are 900 pre-defined queries, and keys and values are the encoder output BEV queries. So the attention for the decoder becomes regular 2D deformable attention~\cite{zhu2020deformable}. Network parameters like eight heads and 0.1 dropout rate are identical to those in the encoder. 

\noindent\textbf{Heatmap branch.} Before we feed the encoder output into the decoder, we first pass it through the heatmap module, which is similar to that in the FCOS3D~\cite{wang2021fcos3d}. The heatmap module is composed of simple convolutional layers, in order to downscale the 256 dimensions of the original BEV queries into a one-dimensional heatmap. As mentioned in the paper, centerness~\cite{zhou2019objects} value ranges from 0 to 1. During the training period, the centerness is 2D Gaussian distribution is written as:
\begin{equation}
    \centering
    c = e^{-\alpha ((\Delta x)^2+(\Delta y)^2)}
    \vspace{-1ex}
\end{equation}
Here, $c$ is the centerness~\cite{zhou2019objects} value. $\alpha$ adjusts the intensity attenuation from the center to the periphery, we set it to $2.5$ for our experiments. Binary Cross Entropy (BCE) loss is used for its supervision, denoted as $\mathcal{L}_{\text{c}}$.

\noindent\textbf{Head and loss.}
We adopt a DETR~\cite{zhu2020deformable} style detection head after the decoder implemented by DETR3D~\cite{wang2022detr3d}, which consists of a classification head and a 3D bounding box head. The classification head produces the logit number of each category. We use Focal loss~\cite{lin2017focal} with $\gamma=2$ and $\alpha=0.25$ to supervise it, denoted as $\mathcal{L}_{\text{cls}}$. Besides, the 3D bounding box head predicts the boxes with the following coefficients: $xyz$ coordinates of the center objects, the length, width, and height of the bounding boxes, the orientation of the boxes and the $xy$ velocity of the objects. It adopts L1 loss $\mathcal{L}_{\text{bbox3d}}$ for 3D bounding box regression. Here we have to notice these two losses have been implemented twice in the network. One of them is after the decoder as the final loss $\mathcal{L}_{\text{bbox3d}}$ which consists of the three losses in total:
\begin{equation}
    \centering
    \mathcal{L} = \lambda_{c}\mathcal{L}_{\text{c}} + \lambda_{cls} \mathcal{L}_{\text{cls}} + \lambda_{bbox3d}\mathcal{L}_{\text{bbox3d}}
    \vspace{-1ex}
\end{equation}
Here we set $\lambda_{cls}=2.0$ and $\lambda_{bbox3d}=0.5$ following the DETR3D~\cite{wang2022detr3d} and BEVFormer~\cite{li2022bevformer}. For the module we added, we make $\lambda_c= 1$ in practice. The other is that after the decoder, using a 3D Hungarian assigner to match the objects and decoder queries. To be more precise, the loss here should be called assigning cost. 

\subsection{Training Setting}
\label{sec:sup_B3_Training Setting}
We provide all the hyperparameters in our model in Table~\ref{tab:sup_t2_training_set}. Nearly all are the same as BEVFormer~\cite{li2022bevformer} and DETR3D~\cite{wang2022detr3d}. To be specific, the first part is about some basic training parameters such as batchsize and epochs while the second part is about the optimizer. We use AdamW with an initial learning rate of 2e-4, and the schedule is Cosine Annealing Warm Restart. Also, we list some other parameters. The image size was set to $1600 \times 900$ for the main experimental results while $1280 \times 720$ for the ablation study. To get historical information, we take the previous four frames for the main experiments. For the ablation experiments, we take the preceding three frames. We illustrate how we use them for temporal modeling in Sec.~\ref{sec:sup_B4_Dataset Pre-processing}.
\begin{table}
    \centering
    \vspace{0pt}
    \begin{tabular}{l|c}
\toprule
\multicolumn{1}{c|}{\bfseries Hyperparameters\mdseries} & \textbf{Value/Type} \\
\midrule

batchsize & 1 \\

training epochs & 24 \\

\midrule

optimizer & AdamW \\

learning rate & 2e-4 \\

weight decay & 0.01 \\

lr schedule & cosine annealing \\

warmup type & linear \\

warmup iters & 500 \\

warmup ratio & 1/3 \\

min lr ratio & 1e-3 \\

gradient clip & 35 \\

\midrule

image size(main) & $1600 \times 900$ \\

image size(ablation) & $1280 \times 720$ \\

queue length(main) & 4 \\

queue length(ablation) & 3 \\

\bottomrule
\end{tabular}
    \vspace{-2mm}
    \caption{Training settings and hyperparameters.} 
    \label{tab:sup_t2_training_set}
\end{table}

\subsection{Dataset Pre-processing}
\label{sec:sup_B4_Dataset Pre-processing}

\noindent\textbf{How we use previous frames for temporal modeling.}
As mentioned earlier, we used the information from the previous frames to model the time. We model the temporal information slightly like the CLIP~\cite{radford2021learning}, which contains several sampling frames, and The last frame corresponds to the current frame. So the first thing is to process the annotation information of the original nuScenes~\cite{caesar2020nuscenes} so that the annotation of each frame contains the previous frames' annotation as well. Similar to the BEVFormer~\cite{li2022bevformer}, we use the 'Can bus' information to record the ego-motion parameters in previous frames. The 'Can bus' in the nuScenes~\cite{caesar2020nuscenes} initially records the states of the ego-motion vehicle, including position, velocity, acceleration, steering, lights, battery, and so on. In temporal modeling, we only consider the position of the ego-motion vehicle. We set the position of the first frame of the sequence to zero, and the position of the subsequent frames to the distance moved relative to the first frame. This operation is offline and is a pre-processing of the overall nuScenes~\cite{caesar2020nuscenes}. In the training dataloader, we save the various camera parameters of the historical frames and the state of the moving objects in them. Moreover, we align the history information between adjacent frames in the sequence. In the test mode, we test one scene like an extensive sequence and align them between adjacent frames.

\noindent\textbf{Pre-processing and Transform.}
We have introduced the offline pre-processing of the entire nuScenes dataset~\cite{caesar2020nuscenes} and the way to align and fuse historical frames. Now we introduce transform operations in the dataloader. In the training pipeline, Photo Metric Distortion was used for data augmentation and we also padded the multi-view images. Also we should note the image size is $1600 \times 900$ for the full version and $80\%$ for the light version in some ablation studies.

\section{Extended Experiment Results}
\label{sec:sup_C_extended_experiment}
In this section, we perform a series of additional experiments on top of the original ones. Here there are two main categories, and in Sec.~\ref{sec:sup_C1_Analysis of Categories in Main Results}. we present a specific analysis of each category for the main experimental results. In Sec.~\ref{sec:sup_C2_Additional Ablation Studies}, we supplement the other ablation experiments.

\subsection{Analysis of Categories in Main Results}
\label{sec:sup_C1_Analysis of Categories in Main Results}

\begin{table*}[t!]\scriptsize
\begin{center}
    \begin{tabular}{l |c c c c c c}
\toprule
Object Class & AP(Precision)~$\uparrow$  & ATE(Translation)~$\downarrow$ & ASE(Scale)~$\downarrow$ & AOE(Orientation)~$\downarrow$ & AVE(Velocity)~$\downarrow$ & AAE(Attribute)~$\downarrow$    \\
\midrule

\footnotesize{car} & 0.617 / 0.621 \better{\betterthanleft{+0.004}} & 0.463 / 0.446 \better{\betterthanleft{-0.017}} & 0.152 / 0.149 \better{\betterthanleft{-0.003}} & 0.067 / 0.057 \better{\betterthanleft{-0.010}} & 0.324 / 0.295 \better{\betterthanleft{-0.029}} & 0.196 / 0.194 \better{\betterthanleft{-0.029}} \\

\footnotesize{truck} & 0.370 / 0.345 \worse{\worsethanleft{-0.025}} & 0.726 / 0.697 \better{\betterthanleft{-0.029}} & 0.212 / 0.204 \better{\betterthanleft{-0.008}} & 0.094 / 0.079 \better{\betterthanleft{-0.015}} & 0.341 / 0.327 \better{\betterthanleft{-0.014}} & 0.194 / 0.195 \worse{\worsethanleft{+0.001}} \\

\footnotesize{bus} & 0.445 / 0.427 \worse{\worsethanleft{-0.018}} & 0.747 / 0.729 \better{\betterthanleft{-0.018}} & 0.212 / 0.202 \better{\betterthanleft{-0.010}} & 0.097 / 0.107 \worse{\worsethanleft{+0.010}} & 0.864 / 0.636 \better{\betterthanleft{-0.228}} & 0.273 / 0.296 \worse{\worsethanleft{+0.023}} \\

\footnotesize{trailer} & 0.171 / 0.168 \worse{\worsethanleft{-0.003}} & 0.971 / 0.929 \better{\betterthanleft{-0.042}} & 0.259 / 0.251 \better{\betterthanleft{-0.008}} & 0.426 / 0.485 \worse{\worsethanleft{+0.059}} & 0.338 / 0.289 \better{\betterthanleft{-0.049}} & 0.078 / 0.065 \better{\betterthanleft{-0.013}} \\

\footnotesize{construction vehicle} & 0.129 / 0.115 \worse{\worsethanleft{-0.019}} & 0.999/ 0.952 \better{\betterthanleft{-0.047}} & 0.451 / 0.455 \worse{\worsethanleft{+0.004}} & 1.157 / 0.954 \better{\betterthanleft{-0.203}} & 0.131 / 0.133 \worse{\worsethanleft{+0.002}} & 0.373 / 0.340 \better{\betterthanleft{-0.034}} \\

\footnotesize{pedestrian} & 0.494 / 0.502 \worse{\worsethanleft{-0.008}} & 0.642 / 0.624 \better{\betterthanleft{-0.018}} & 0.294 / 0.296 \worse{\worsethanleft{+0.002}} & 0.432 / 0.410 \better{\betterthanleft{-0.022}} & 0.361 / 0.345 \better{\betterthanleft{-0.016}} & 0.174 / 0.157 \better{\betterthanleft{-0.017}} \\

\footnotesize{motorcycle} & 0.433 / 0.434 \better{\betterthanleft{+0.001}} & 0.643 / 0.589 \better{\betterthanleft{-0.054}} & 0.257 / 0.261 \worse{\worsethanleft{+0.004}} & 0.441 / 0.342 \better{\betterthanleft{-0.099}} & 0.441 / 0.466 \worse{\worsethanleft{+0.025}} & 0.266 / 0.232 \better{\betterthanleft{-0.034}} \\

\footnotesize{bicycle} & 0.399 / 0.396 \worse{\worsethanleft{-0.003}} & 0.555 / 0.541 \better{\betterthanleft{-0.014}} & 0.272 / 0.281 \worse{\worsethanleft{+0.009}} & 0.485 / 0.516 \worse{\worsethanleft{+0.031}} & 0.253 / 0.246 \better{\betterthanleft{-0.007}} & 0.025 / 0.013 \better{\betterthanleft{-0.012}} \\

\footnotesize{traffic cone} & 0.584 / 0.615 \better{\betterthanleft{+0.031}} & 0.452 / 0.372 \better{\betterthanleft{-0.080}} & 0.331 / 0.332 \worse{\worsethanleft{+0.001}} & - / - & - / - & - / - \\

\footnotesize{barrier} & 0.524 / 0.551 \better{\betterthanleft{+0.027}} & 0.531 / 0.414 \better{\betterthanleft{-0.117}} & 0.292 / 0.299 \worse{\worsethanleft{+0.007}} & 0.139 / 0.101 \better{\betterthanleft{-0.038}} & - / - & - / - \\

\bottomrule
\end{tabular}
    \vspace{-5pt}
    \caption{\textbf{Categories details in the Main Result.} Here we show the main results comparison of BEVFormer~\cite{li2022bevformer} and our OCBEV. The data on the left of the slash is the result of BEVFormer~\cite{li2022bevformer}, and the right is that of OCBEV. Data on the far right is the difference. Green means that OCBEV is better than BEVFormer~\cite{li2022bevformer}, and gray means OCBEV is worse.}
    \vspace{-12pt}
    \label{tab:sup_t3_categories_in_main_results}
\end{center}
\subfloat[
\label{tab:sup_t4a_add_ablation}
\textbf{Loss Weight.} The loss weight is weights of classification and 3D bounding boxes and we set the centerness loss as 1 in practice. Its fluctuation range becomes $0.2\%$ in NDS and $0.4\%$ in mAP which means that the effect of the loss weight on the main experimental results is not significant. 
]
{
\centering
\begin{minipage}{0.31\linewidth}{\begin{center}
\tablestyle{4pt}{1.10}
\begin{tabular}{l |c|c c}
\toprule
Backbone  & \textbf{Loss Weight} & NDS~$\uparrow$  & mAP~$\uparrow$ \\
\midrule
ResNet-101 & 2 : 0.4 & 0.528 & 0.413  \\
\rowcolor{bar!10}
ResNet-101 & 2 : 0.5 & 0.532 & 0.417  \\
ResNet-101 & 2 : 0.6 & 0.531 & 0.414 \\
ResNet-101 & 2 : 0.7 & 0.530 & 0.413 \\

\bottomrule
\end{tabular}
\end{center}}\end{minipage}
}
\hspace{2mm}
\subfloat[
\textbf{Learning Rate.} We change the learning rate for the whole network. All learning rates are in the $10^{-4}$ level. Its fluctuation range becomes $4.5\%$ in NDS and $10.1\%$ in mAP which means that the effect of the learning rate on the main experimental results is extremely significant.  
\label{tab:sup_t4b_add_ablation}
]
{
\begin{minipage}{0.31\linewidth}{\begin{center}
\tablestyle{4pt}{1.10}
\begin{tabular}{l |c |c c }
\toprule
Backbone & \textbf{LR} & NDS~$\uparrow$  & mAP~$\uparrow$ \\
\midrule
ResNet-101 & $1\times10^{-4}$ & 0.509 & 0.316 \\
\rowcolor{bar!10}
ResNet-101 & $2\times10^{-4}$ & 0.532 & 0.417 \\
ResNet-101 & $4\times10^{-4}$ & 0.487 & 0.375 \\
ResNet-101 & $6\times10^{-4}$ & 0.493 & 0.356 \\

\bottomrule
\end{tabular}
\end{center}}\end{minipage}
}
\hspace{4mm}
\subfloat[
\label{tab:sup_t4c_add_ablation}
\textbf{BEV Size.} BEV size below 200 has been explored in BEVFormer so We set the BEV size larger than 200 respectively. Its fluctuation range becomes $0.4\%$ in NDS and $0.3\%$ in mAP which means that Our model is very robust to the parameter BEV size and its effect is very slight.
]
{
\centering
\begin{minipage}{0.31\linewidth}{\begin{center}
\tablestyle{4pt}{1.10}

\begin{tabular}{l |c|c c}
\toprule
Backbone  & \textbf{BEV Size} & NDS~$\uparrow$  & mAP~$\uparrow$ \\
\midrule
ResNet-101 & $200\times200$ & 0.531 & 0.414  \\
ResNet-101 & $300\times300$ & 0.532 & 0.417  \\
ResNet-101 & $350\times350$ & 0.529 & 0.415 \\
ResNet-101 & $400\times400$ & 0.528 & 0.415 \\

\bottomrule
\end{tabular}
\end{center}}\end{minipage}
}
\vspace{-5pt}
\caption{
\textbf{Ablation study.} We conducted additional ablation experiments on a full version OCBEV model trained with 12 epochs and the configuration of the model is the same as that in the main result experiments. Our results indicate that the model is robust to \textbf{loss weight} and \textbf{BEV size}, and \textbf{learning rate} have a significant impact on the performance of the model.
}
\label{tab:sup_t4_add_ablation} 
\vspace{-10pt}
\end{table*}
The most complete experimental configuration was used for the main experimental results for comparison with other methods. Here we listed the results of each category. There are 10 evaluable categories in total as shown in Table~\ref{tab:sup_t3_categories_in_main_results}. For each one, we show all its six metrics, including precision, translation, scale, orientation, velocity and attribute. For each small pane, there are three pieces of data. The data to the left of the slash is the BEVFormer~\cite{li2022bevformer} result, the data to the right of the slash is our OCBEV result, and the data in the right column is the result obtained by subtracting the BEVFormer~\cite{li2022bevformer} result from the OCBEV result. The green bar means that OCBEV is better than BEVFormer~\cite{li2022bevformer}, and the gray bar means ours is worse than BEVFormer~\cite{li2022bevformer}.

In terms of precision, BEVFormer~\cite{li2022bevformer} and OCBEV are almost the same, OCBEV performs poorly on  large objects such as truck, bus and trailer and performs well on normal moving objects like car and motocycle. Due to OCBEV is Object-Centric, OCBEV is particularly prominent in metrics about the state of motion of objects like translation, orientation and velocity, which means our Object-centric module model the motion characteristics of objects better. Also OCBEV is also better than BEVFormer~\cite{li2022bevformer} in terms of the attribute metric, but slightly worse in terms of scale. From the perspective of each category, OCBEV outperforms BEVFormer~\cite{li2022bevformer} in all aspects for the car category and has better performance in other kinds of objects as well.

\subsection{Additional Ablation Studies}
\label{sec:sup_C2_Additional Ablation Studies}
A series of additional experiments were carried out for the main results to exclude the interference of other factors. The setting of ablation experiments here differs from that of the main paper as the model configurations here are the same as those in the main results \ie 6 layers of encoder and decoder. We did not disable the temporal attention as well.

\noindent\textbf{Loss weight.} We do additional ablation experiments on the loss weitht, learning rate and BEV size to exclude these factors. About the loss weights, the loss we added $\mathcal{L}_{\text{c}}$, and we set its weight $\lambda_{c}=1$ as practice. But for the loss weights of classification and bounding boxes, we follow the DETR3D~\cite{wang2022detr3d} and BEVFormer~\cite{li2022bevformer}, and do ablation experiments on the loss weight. Table~\ref{tab:sup_t4a_add_ablation} shows that the effect of loss weight on the experimental results is not significant. 

\noindent\textbf{Learning rate.} We also follow the DETR3D~\cite{wang2022detr3d} and BEVFormer~\cite{li2022bevformer} and set the learning rate to $2 \times 10^{-4}$. Here we explore the effect of the learning rate on the experiments as well. Table~\ref{tab:sup_t4b_add_ablation} shows that the learning rate has a significant effect on the results. The fluctuation range reaches $4.5\%$ in NDS ($48.7\%$ and $53.2\%$) and $10.1\%$ in the mAP metric.
\vspace{1pt}

\noindent\textbf{BEV size.} As mentioned in the main paper. We did not follow the BEV size in DETR3D~\cite{wang2022detr3d} and BEVFormer~\cite{li2022bevformer} which is $200\times200$ and we set the batchsize to $300\times300$.  We also performed additional ablation experiments for BEV size. BEV size below 200 has been explored in BEVFormer~\cite{li2022bevformer} so We set it larger than 200. Table~\ref{tab:sup_t4c_add_ablation} shows that BEV size is an irrelevant variable.

\section{Visualization}
\label{sec:sup_D_visualization}
We demonstrate the visualization and qualitative results of our 3D object detection in Fig.~\ref{fig:sup_f1}. We give the qualitive results of perspective views and the BEV view. We compare our results with BEVFormer. It is shown that our OCBEV predicts more accurate bounding boxes than that of BEVFormer. OCBEV can detect hard cases in the distance or with occlusion which prove OCBEV is more advanced.

\begin{figure*}[tbp]
\centering
\subfloat[OVBEV can reduce false positive cases. BEVFormer predicts a car in front view but actually there is no car there at all.]{
\label{fig:sup_f4a_visualization1}
\includegraphics[width=0.99\linewidth]{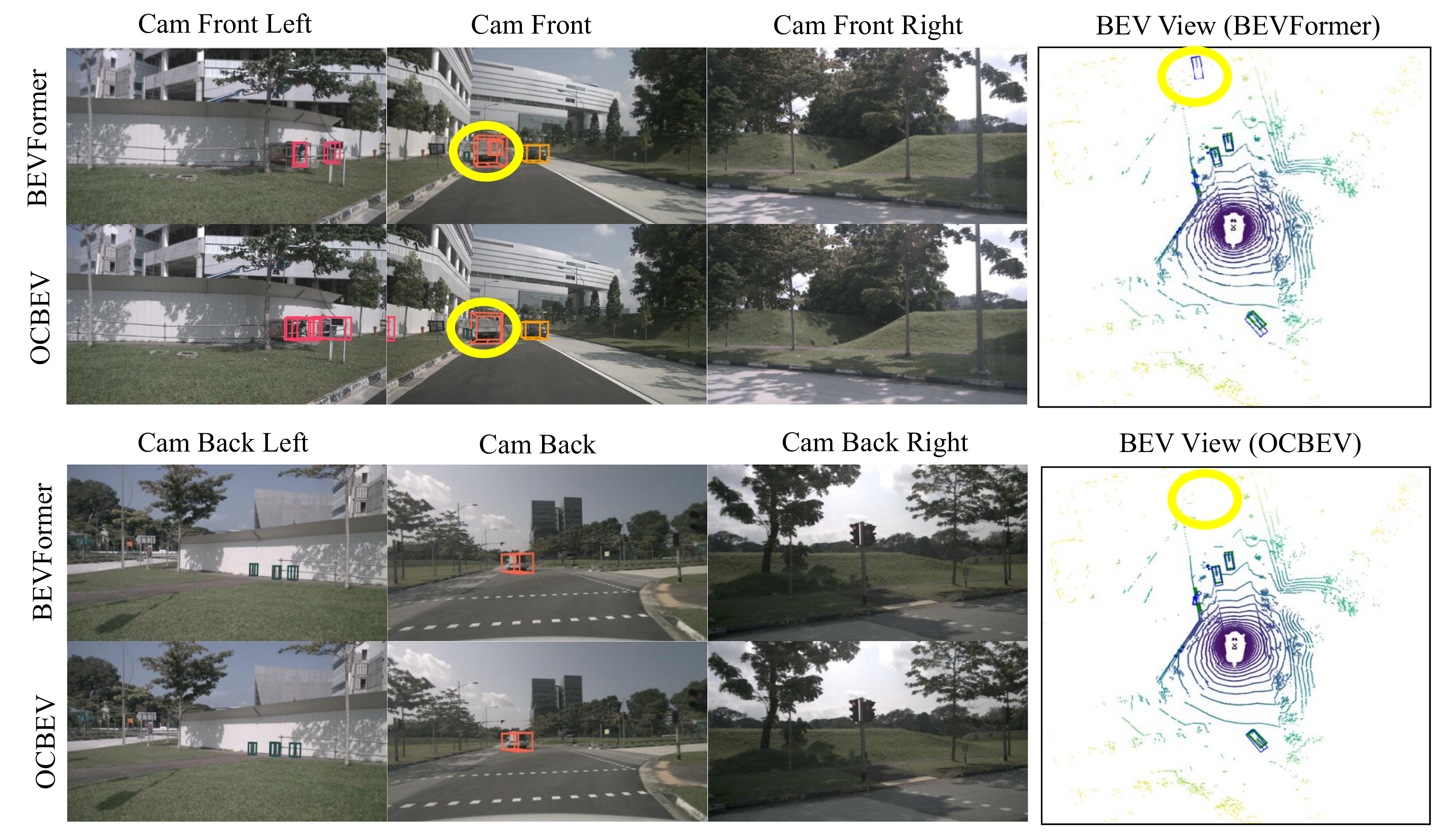}}\\
\vspace{2mm}
\subfloat[OCBEV can give more precise locations and poses . BEVFormer predicts locations and poses of cars in back left view poorly than OCBEV.]{
\label{fig:sup_f4b_visualization2}\includegraphics[width=0.99\linewidth]{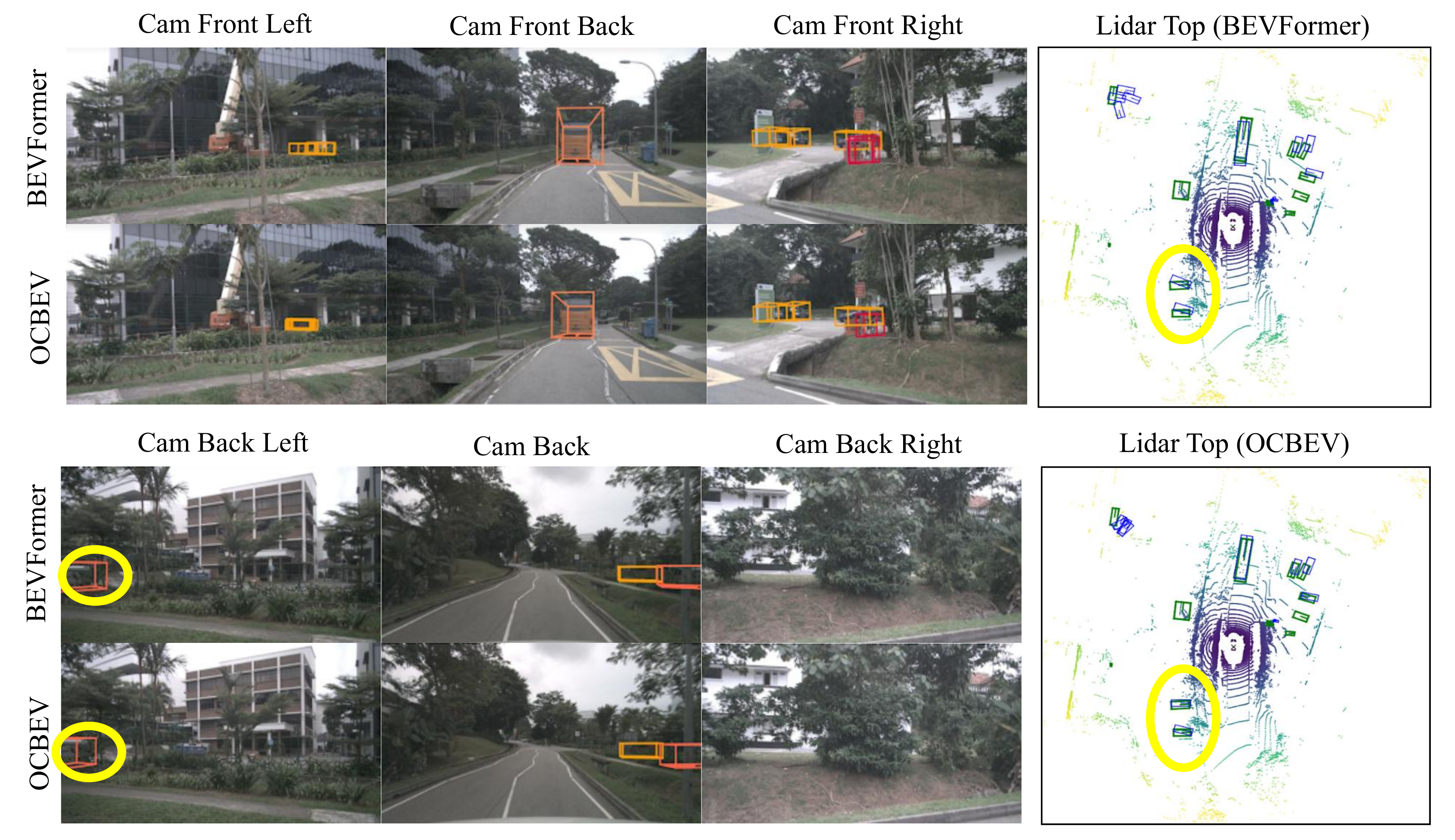}} \\
\caption{\textbf{Comparision of BEVFormer and our OCBEV on the nuScenes val set.} In BEV view, the green boxes are the ground truth and blue boxes are predictions. And OCBEV can reduce false positive cases and give more accuracy boxes}
\label{fig:sup_f4}
\end{figure*}

\begin{figure*}[tbp]
\centering
\subfloat[OCBEV detect the Fedex car in the front view while BEVFormer does not detect that sucessfully.]{
\label{fig:sup_f5a_visualization3}
\includegraphics[width=0.99\linewidth]{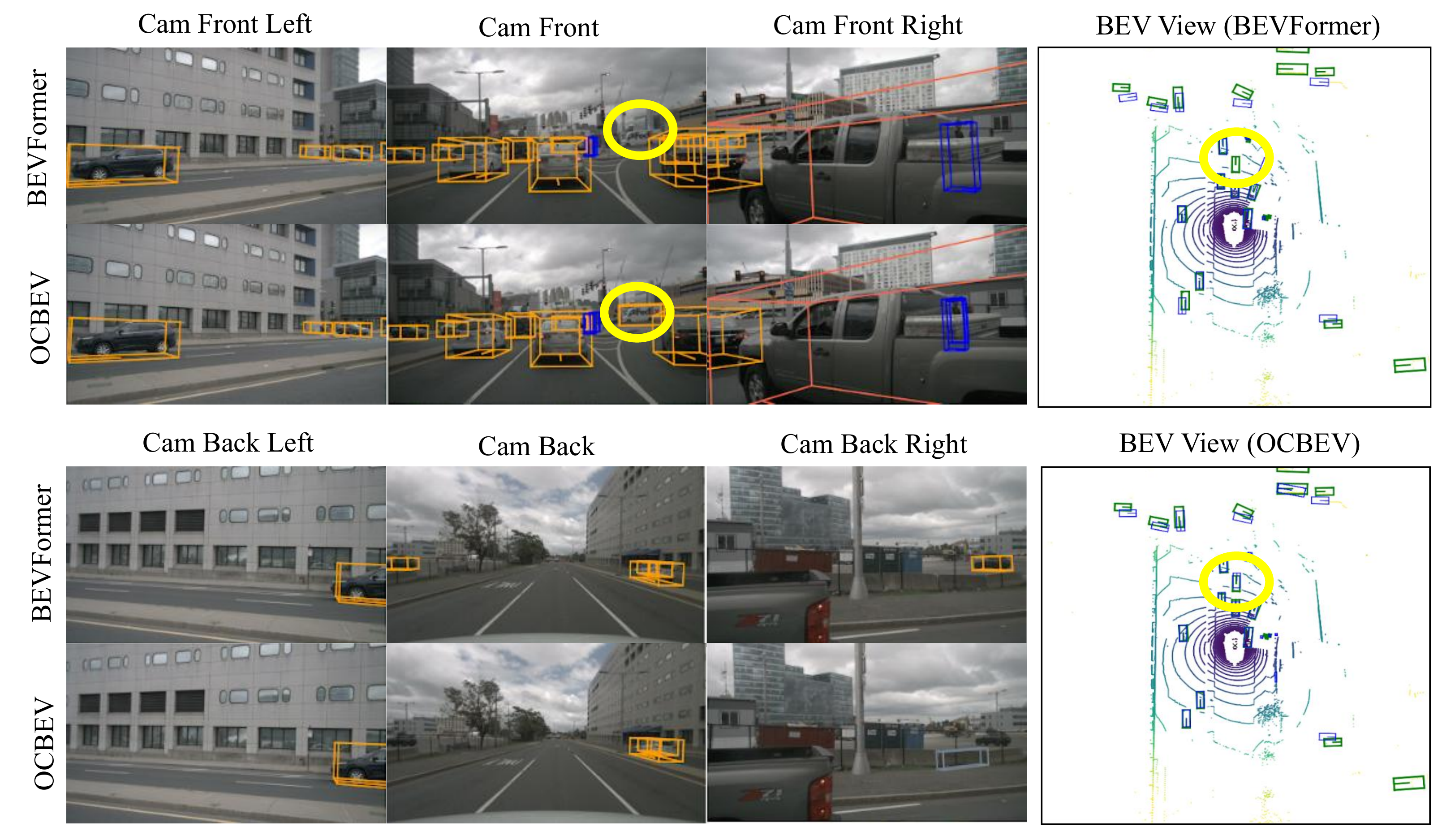}}\\
\vspace{2mm}
\subfloat[For dense vehicle sceneslike car park, OCBEV can give better performance than BEVFormer.]{
\label{fig:sup_f5b_visualization4}\includegraphics[width=0.99\linewidth]{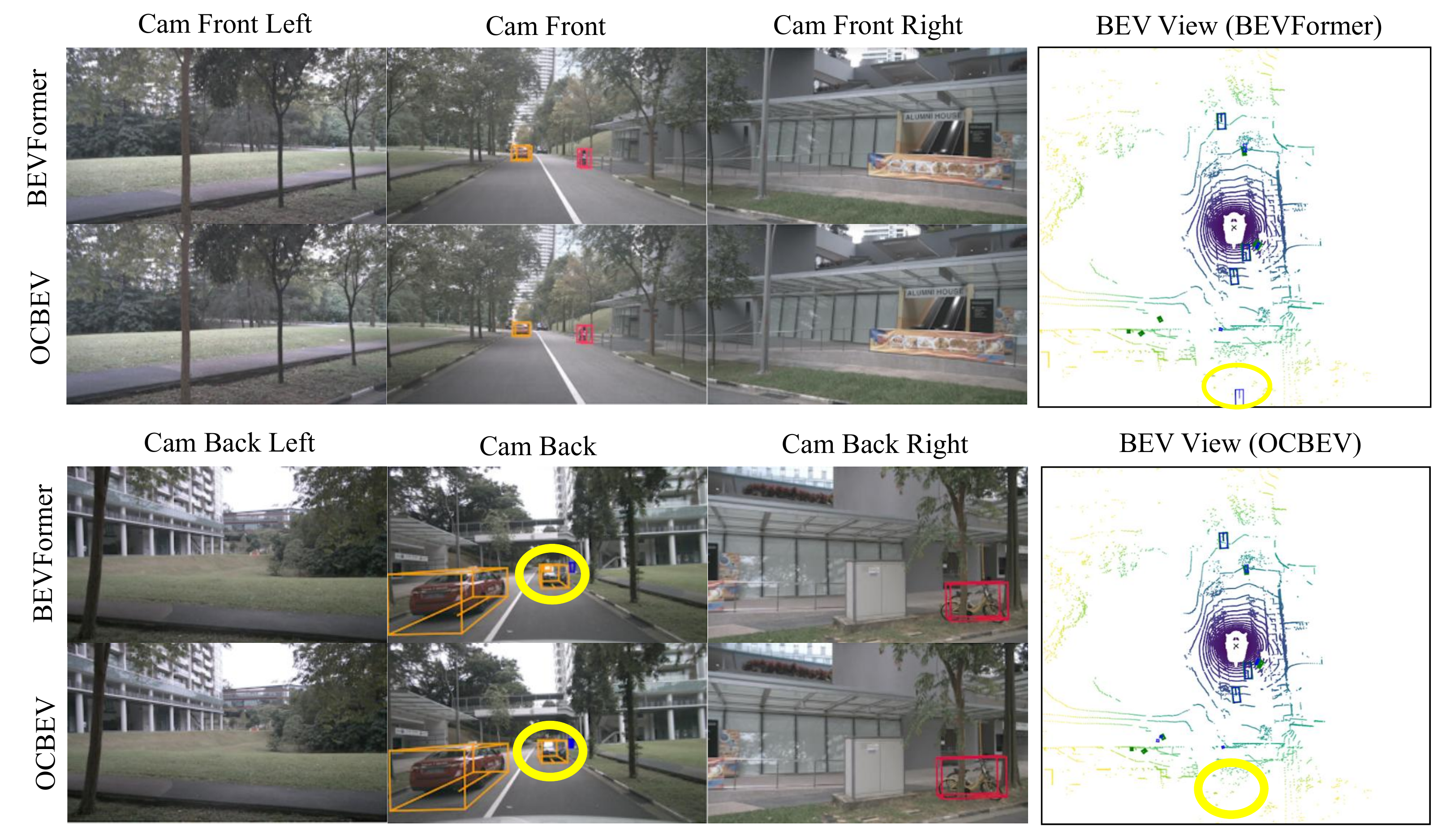}} \\
\caption{\textbf{Comparision of BEVFormer and our OCBEV on the nuScenes val set.} In BEV view, the green boxes are the ground truth and blue boxes are predictions. OCBEV have higher precision and can handle dense vehicle scenes.}
\label{fig:sup_f5}
\end{figure*}

\end{document}